\documentclass[11pt,a4paper]{article}
\usepackage[hyperref]{emnlp2020}
\usepackage{times}
\usepackage{latexsym}

\usepackage{microtype}

\aclfinalcopy 
\def\aclpaperid{319} 

\usepackage{url}
\usepackage{tabulary}
\usepackage{tabularx}
\usepackage[normalem]{ulem}
\usepackage{enumitem}
\usepackage{multirow}
\usepackage{array}
\usepackage{booktabs}
\usepackage{graphicx}
\usepackage{subfigure}
\usepackage{makecell}
\usepackage{amsmath,amsfonts,amsthm,amssymb,amsopn,bm}
\usepackage{color,soul}
\usepackage{verbatim}
\usepackage{pifont}
\usepackage{footnote}
\DeclareMathOperator*{\argmax}{argmax}
\DeclareMathOperator*{\argmin}{argmin}
\DeclareMathOperator*{\topk}{topK}

\newcommand{\roberta}{\textsc{RoBERTa}}
\newcommand{\robertalarge}{\ensuremath{\textsc{RoBERTa}_{\textsc{LARGE}}}}
\newcommand{\bert}{\textsc{BERT}}
\newcommand{\bertbase}{\ensuremath{\textsc{BERT}_{\textsc{BASE}}}}
\newcommand{\bertlarge}{\ensuremath{\textsc{BERT}_{\textsc{LARGE}}}}
\newcommand{\hotpot}{\textsc{HotpotQA}}
\newcommand{\squad}{\textsc{SQuAD}}
\newcommand{\decomprc}{\textsc{DecompRC}}
\newcommand{\yes}{\texttt{yes}}
\newcommand{\no}{\texttt{no}}

\newcommand{\xmark}{\ding{55}}
\newcommand\ignore[1]{}

\newcommand{\cseqtoseq}{CONUS}
\newcommand{\useqtoseq}{ONUS}
\newcommand{\seqtoseq}{Seq2Seq}

\title{Unsupervised Question Decomposition for Question Answering}

\author{Ethan Perez$^{1~2}$ ~~ Patrick Lewis$^{1~3}$\\\textbf{Wen-tau Yih}$^1$ ~~ \textbf{Kyunghyun Cho}$^{2~4}$\thanks{~~KC was a part-time research scientist at Facebook AI Research while working on this paper.} ~~ \textbf{Douwe Kiela}$^1$\\
$^1$Facebook AI Research, $^2$New York University,\\$^3$University College London, $^4$CIFAR Azrieli Global Scholar\\
  {\tt perez@nyu.edu} \\}

\date{}

\begin{document}
\maketitle
\begin{abstract}
We aim to improve question answering (QA) by decomposing hard questions into simpler sub-questions that existing QA systems are capable of answering. Since labeling questions with decompositions is cumbersome, we take an unsupervised approach to produce sub-questions, also enabling us to leverage millions of questions from the internet. Specifically, we propose an algorithm for One-to-N Unsupervised Sequence transduction (\useqtoseq) that learns to map one hard, multi-hop question to many simpler, single-hop sub-questions. We answer sub-questions with an off-the-shelf QA model and give the resulting answers to a recomposition model that combines them into a final answer. We show large QA improvements on \hotpot{} over a strong baseline on the original, out-of-domain, and multi-hop dev sets. \useqtoseq{} automatically learns to decompose different kinds of questions, while matching the utility of supervised and heuristic decomposition methods for QA and exceeding those methods in fluency. Qualitatively, we find that using sub-questions is promising for shedding light on why a QA system makes a prediction.\footnote{Our code, data, and pretrained models are available at \url{https://github.com/facebookresearch/UnsupervisedDecomposition}.}
\end{abstract}

\section{Introduction}
\label{sec:Introduction}


\begin{figure}[t]
\centering
\includegraphics[width=0.99\columnwidth]{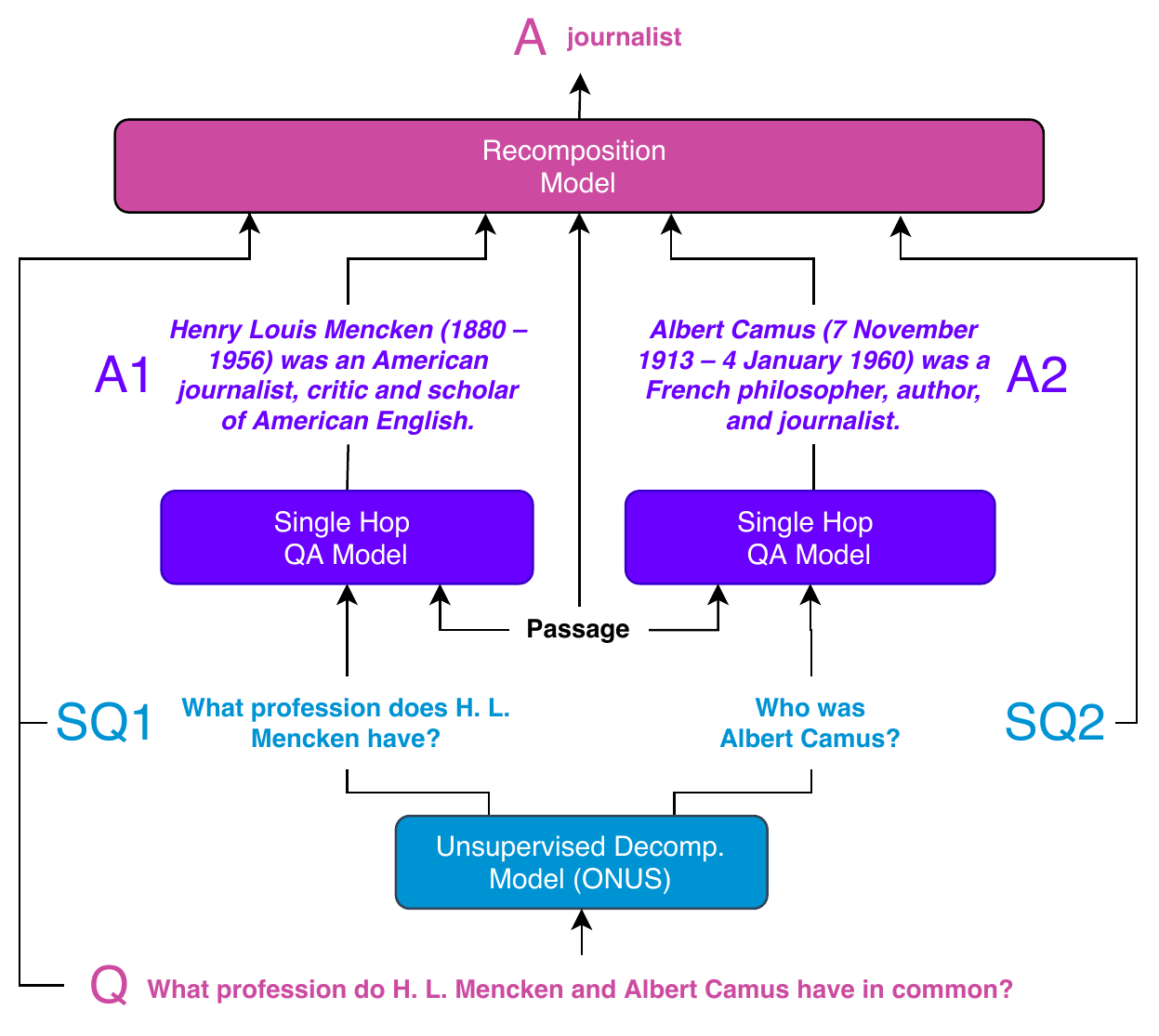}
\caption{\textit{Overview}: Using unsupervised learning, we decompose a multi-hop question into single-hop sub-questions, whose predicted answers are given to a recomposition model to predict the final answer.}
\label{fig:decomposition_for_qa}
\end{figure}


It has been a long-standing challenge in AI to answer questions of any level of difficulty~\cite{winograd1991thinking}.
Question answering (QA) systems struggle to answer complex questions such as \textit{``What profession do H. L. Mencken and Albert Camus have in common?''} since the required information is scattered in different places~\cite{yang2018hotpotqa}. However, QA systems accurately answer simpler, related questions such as \textit{``What profession does H. L. Mencken have?''} and \textit{``Who was Albert Camus?''}~\cite{petrochuk2018simplequestions}.
Thus, a promising strategy to answer hard questions is divide-and-conquer: decompose a hard question into simpler sub-questions, answer the sub-questions with a QA system, and recompose the resulting answers into a final answer, as shown in Figure~\ref{fig:decomposition_for_qa}.
This approach leverages strong performance on simple questions to help answer harder questions~\cite{christiano2018supervising}.

Existing work decomposes questions using a combination of hand-crafted heuristics, rule-based algorithms, and learning from supervised decompositions~\cite{talmor2018web,min2019multi}, which each require significant human effort.
For example, \decomprc{}~\cite{min2019multi} decomposes some questions using supervision and other questions using a heuristic algorithm with fine-grained, special case handling based on part-of-speech tags and over 50 keywords.
Prior work also assumes that sub-questions only consist of words from the question, which is not always true.
Decomposing arbitrary questions requires sophisticated natural language generation, which often relies on many, high-quality supervised examples.
Instead of using supervision, we find it possible to decompose questions in a \textit{fully unsupervised} way.

\begin{figure}[t]
\centering
\includegraphics[width=\columnwidth]{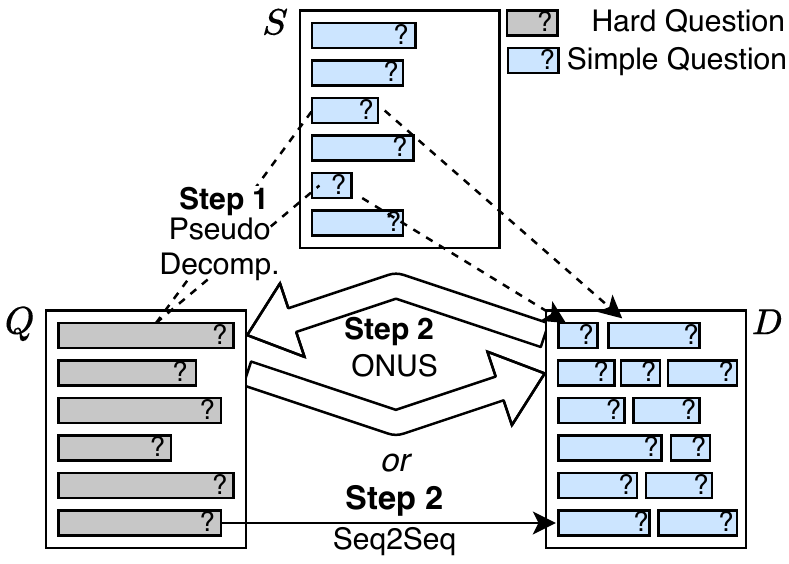}
\caption{\textit{One-to-N Unsupervised Sequence transduction (ONUS)}: \textbf{Step 1}: We create a corpus of pseudo-decompositions $D$ by finding candidate sub-questions from a simple question corpus $S$ which are similar to a multi-hop question in $Q$. \textbf{Step 2}: We learn to map multi-hop questions to decompositions using $Q$ and $D$ as training data, via either standard sequence-to-sequence learning (\seqtoseq) or unsupervised sequence-to-sequence learning (for \useqtoseq).}
\label{fig:unsupervised_decomposition}
\end{figure}

We propose an algorithm for One-to-N Unsupervised Sequence transduction (\useqtoseq) that learns to map from the distribution of hard questions to that of \textit{many} simple questions.
First, we automatically create a noisy ``pseudo-decomposition'' for each hard question by using embedding similarity to retrieve sub-question candidates.
We mine over 10M possible sub-questions from Common Crawl with a classifier, showcasing the effectiveness of parallel corpus mining, a common approach in machine translation~\cite{xu2017zipporah,artetxe2019margin}, for QA.
Second, we train a decomposition model on the mined data with unsupervised sequence-to-sequence learning, allowing \useqtoseq{} to improve over pseudo-decompositions.
As a result, we are able to train a large transformer model to generate decompositions, surpassing the fluency of heuristic/extractive decompositions.
Figure~\ref{fig:unsupervised_decomposition} overviews our approach to decomposition.

We validate \useqtoseq{} on multi-hop QA, where questions require reasoning over multiple pieces of evidence.
We use an off-the-shelf single-hop QA model to answer decomposed sub-questions.
Then, we give sub-questions and their answers to a recomposition model to combine into a final answer.
We evaluate on three dev sets for \hotpot{}, a standard benchmark for multi-hop QA~\cite{yang2018hotpotqa}, including two challenge sets.

\useqtoseq{} proves to be a powerful tool for QA in the following ways.
First, QA models that use decompositions outperform a strong RoBERTa baseline~\cite{liu2019roberta,min2019compositional} by 3.1 points in F1 on the original dev set, 10 points on the out-of-domain dev set from~\citet{min2019multi}, and 11 points on the multi-hop dev set from~\citet{jiang2019avoiding}.
Our method is competitive with state-of-the-art methods SAE~\cite{tu2020select} and HGN~\cite{fang2019hierarchical} that use additional, strong supervision on which sentences are relevant to the question.
Second, our analysis shows that sub-questions improve multi-hop QA by using the single-hop QA model to retrieve question-relevant text.
Qualitative examples illustrate how the retrieved text adds a level of interpretability to otherwise black-box, neural QA models.
Third, \useqtoseq{} automatically learns to generate useful decompositions for all four question types in \hotpot, highlighting the general nature of \useqtoseq{} over prior work, such as IBM Watson~\cite{ferrucci2010building} and \decomprc{}~\cite{min2019multi}, which decompose different question types separately.
Without finetuning, our trained \useqtoseq{} model can even decompose some questions in visual QA~\cite{johnson2017clevr} and knowledge-base QA~\cite{talmor2018web}, as well as claims in fact verification~\cite{thorne2018fever}, suggesting promising future avenues in other domains.

\section{Method}
\label{sec:Method}
We now formulate the problem and describe our high-level approach, with further details in \S\ref{sec:Experimental Setup}.
The goal of this work is to leverage a QA model that is accurate on simple questions for answering hard questions, without using annotated question decompositions.
Here, we consider simple questions to be ``single-hop'' questions that require reasoning over one paragraph or piece of evidence, and we consider hard questions to be ``multi-hop.''
Our aim is to train a multi-hop QA model $M$ to provide the correct answer $a$ to a multi-hop question $q$ about a given context $c$ (e.g., several paragraphs). Normally, we would train $M$ to maximize $\log p_M(a | c, q)$. To facilitate learning, we leverage a single-hop QA model that may be queried with sub-questions $s_1, \dots, s_N$, whose ``sub-answers'' $a_1, \dots, a_N$ may be given to $M$. $M$ may then maximize the potentially easier objective $\log p_M(a | c, q, [s_1, a_1], \dots, [a_N, s_N])$.

Supervised decomposition models learn to map each question $q \in Q$ to a decomposition $d = [s_1; \dots; s_N]$ of $N$ sub-questions $s_n \in S$ using annotated $(q, d)$ examples.
In this work, we do not assume access to strong $(q, d)$ supervision.
To leverage the single-hop QA model without supervision, we follow a three-stage approach: 1) map a question $q$ into sub-questions $s_1, \dots, s_N$ via unsupervised techniques, 2) find sub-answers $a_1, \dots, a_N$ with the single-hop QA model, and 3) use $s_1, \dots, s_N$ and $a_1, \dots, a_N$ to predict $a$.

\subsection{Unsupervised Question Decomposition}
\label{ssec:Unsupervised Decomposition}
To train an unsupervised decomposition model, we need suitable data.
We assume access to a hard question corpus $Q$ and simple question corpus $S$.
Instead of using supervised $(q, d)$ examples, we design an algorithm that creates pseudo-decompositions $d'$ to form $(q, d')$ pairs from $Q$ and $S$ using an unsupervised method (\S\ref{sssec:Creating Pseudo-Decompositions}).
We then train a model to map $q$ to a decomposition.
We explore learning to decompose with standard and unsupervised sequence-to-sequence learning (\S\ref{sssec:Learning to Decompose}).

\subsubsection{Creating Pseudo-Decompositions}
\label{sssec:Creating Pseudo-Decompositions}
Inspired by~\citet{zhou2015learning} in question retrieval, we create a pseudo-decomposition set $d' = \{s_1; \dots; s_N\}$ for each $q \in Q$ by retrieving simple question $s_i$ from $S$.
We concatenate $s_1; \dots; s_N$ to form $d'$ used downstream.
$N$ may potentially vary based on $q$.
To retrieve useful simple questions for answering $q$, we face a joint optimization problem. We want sub-questions that are both (i) similar to $q$ according to a metric $f$ (first term) and (ii) maximally diverse (second term), so our objective is:
\begin{equation}
\label{eqn:generalized_similarity_retrieval_eqn}
  \argmax_{d' \subset S} \sum_{s_i \in d'} f(q, s_i) - \sum_{\substack{s_i, s_j \in d', i \ne j}} f(s_i, s_j)
\end{equation}

\subsubsection{Learning to Decompose}
\label{sssec:Learning to Decompose}
With the above pseudo-decompositions, we explore various decomposition methods~(details in \S\ref{sssec:Unsupervised Decomposition Models}):

\paragraph{PseudoD}
We use sub-questions from pseudo-decompositions directly in downstream QA.

\paragraph{Sequence-to-Sequence (\seqtoseq)}
We train a Seq2Seq model $p_{\theta}$ to maximize $\log p_{\theta}(d' | q)$.

\paragraph{One-to-N Unsupervised Sequence transduction (\useqtoseq)}
We use unsupervised learning to map one question to $N$ sub-questions.
We start with paired $(q, d')$ but do not learn from the pairing because it is noisy.
Instead, we use unsupervised Seq2Seq methods to learn a $q \rightarrow d$ mapping.

\subsection{Answering Sub-Questions}
\label{ssec:Answering Sub-Questions}
To answer the generated sub-questions, we use an off-the-shelf QA model.
The QA model may answer sub-questions using any free-form text (i.e., a word, phrase, sentence, etc.).
Any QA model is suitable, so long as it can accurately answer simple questions in $S$.
We thus leverage good accuracy on questions in $S$ to help answer questions in $Q$.

\subsection{Learning to Recompose}
\label{ssec:Multi-hop QA using Sub-Questions}
Downstream QA systems may use sub-questions and sub-answers in various ways.
We train a recomposition model to combine the decomposed sub-questions/answers into a final answer, when also given the original input (context+question).

\section{Experimental Setup}
\label{sec:Experimental Setup}
We now detail the implementation of our approach.

\subsection{Question Answering Task}
We test \useqtoseq{} on \hotpot{}, a standard multi-hop QA benchmark.
Questions require information from two distinct Wikipedia paragraphs to answer (\textit{``Who is older, Annie Morton or Terry Richardson?''}).
For each question, \hotpot{} provides 10 context paragraphs from Wikipedia.
Two paragraphs contain question-relevant sentences called ``supporting facts,'' and the remaining paragraphs are irrelevant, ``distractor paragraphs.''
Answers in \hotpot{} are either \yes, \no, or a text span in an input paragraph.
Accuracy is measured with F1 word overlap and Exact Match (EM) between predicted and gold spans.

\subsection{Unsupervised Decomposition}
\subsubsection{Training Data and Question Mining}
\label{ssec:Question Data}
Supervised decomposition methods are limited by the amount of available human annotation, but our unsupervised method faces no such limitation, similar to unsupervised QA~\cite{lewis2019unsupervised}.
Since we need to train data-hungry Seq2Seq models, we would benefit from large training corpora.
A larger simple question corpus will also improve the relevance of retrieved simple questions to the hard question.
Thus, we take inspiration from parallel corpus mining in machine translation~\cite{xu2017zipporah,artetxe2019margin}.
We use questions from \squad{} 2 and \hotpot{} to form our initial corpora $S$ (single-hop questions) and $Q$ (multi-hop questions), respectively, and we augment $Q$ and $S$ by mining more questions from Common Crawl.
First, we select sentences that start with ``wh''-words or end in ``?''
Next, we train an efficient, FastText classifier~\cite{joulin-etal-2017-bag} to classify between questions sampled from Common Crawl, \squad{} 2, and \hotpot{} (60K in total).
Then, we classify our Common Crawl questions, adding those classified as \squad{}~2 questions to $S$ and those classified as \hotpot{} questions to $Q$.
Mining greatly increases the number of single-hop questions (130K $\rightarrow$ 10.1M) and multi-hop questions (90K $\rightarrow$ 2.4M), showing the power of parallel corpus mining in QA.
\footnote{See Appendix \S\ref{ssec:apdx:Question Mining Details} for details on question classifier.}

\subsubsection{Creating Pseudo-Decompositions}
\label{ssec:Creating Pseudo-Decompositions Experimental Setup}
To create pseudo-decompositions (retrieval-based sub-questions for a given question), we experimented with using a variable number of sub-questions $N$ per question (Appendix \S\ref{ssec:apdx:generalized_pseudo_decomposition}), but we found similar QA results with a fixed $N=2$, which we use in the remainder for simplicity.

\paragraph{Similarity-based Retrieval}
To retrieve relevant sub-questions, we embed any text $t$ into a vector $\mathbf{v}_t$ by summing the FastText vectors~\cite{bojanowski-etal-2017-enriching}\footnote{300-dim. English Common Crawl vectors: \url{https://fasttext.cc/docs/en/english-vectors.html}} for words in $t$ and use cosine as our similarity metric $f$.\footnote{We also tried TFIDF and \bert{} representations but did not see improvements over FastText (see Appendix \S\ref{ssec:apdx:pseudo_decomposition_retrieval_method}).}
Let $q$ be a multi-hop question with a pseudo-decomposition $(s_1^*, s_2^*)$ and $\hat{\mathbf{v}}$ be the unit vector of $\mathbf{v}$.
Since $N=2$, Eq. \ref{eqn:generalized_similarity_retrieval_eqn} simplifies to:
\begin{equation*}
     (s_1^*, s_2^*) = \argmax_{\{s_1, s_2 \}\in S}\left[\mathbf{\hat{v}}_q^\top \mathbf{\hat{v}}_{s_1}+ \mathbf{\hat{v}}_q^\top \hat{\mathbf{v}}_{s_2} - \mathbf{\hat{v}}_{s_1}^\top \mathbf{\hat{v}}_{s_2} \right]
\end{equation*}
The last term requires $O(|S|^2)$ comparisons, which is expensive as $|S|>$ 10M.
Instead of solving the above equation exactly, we find an approximate pseudo-decomposition $(s_1', s_2')$ by computing over $S' = \topk_{\{s \in S\}}\left[ \mathbf{\hat{v}}_{q}^{\top} \mathbf{\hat{v}}_s\right]$ with $K=1000$.
We efficiently build $S'$ with FAISS~\cite{JDH17}.

\paragraph{Random Retrieval}
For comparison, we test a random pseudo-decomposition baseline, where we retrieve $s_1, \dots, s_N$ by sampling uniformly from $S$.

\paragraph{Editing Pseudo-Decompositions}
Since sub-questions are retrieval-based, they are often not about the same entities as $q$.
Inspired by retrieve-and-edit methods~\citep[e.g., ][]{guu-etal-2018-generating}, we replace each sub-question entity not in $q$ with an entity from $q$ of the same type (e.g., ``Date'' or ``Location'') if possible.\footnote{Entities found with spaCy~\cite{spacy2}.}
This step is important for PseudoD and \seqtoseq{} (which would learn to hallucinate entities) but not \useqtoseq{} (which must reconstruct entities in $q$ from its own decomposition, as discussed next).

\subsubsection{Unsupervised Decomposition Models}
\label{sssec:Unsupervised Decomposition Models}
\paragraph{Pretraining}
Pretraining is crucial for unsupervised Seq2Seq methods~\cite{artetxe2018unsupervised-neural,lample2018unsupervised}, so we initialize all decomposition models (\seqtoseq{} or \useqtoseq{}) with the same pretrained weights.
We warm-start our pretraining with the pretrained, English Masked Language Model (MLM) from~\citet{lample2019cross}, a 12-block transformer~\citep{vaswani2018attention}.
We do MLM finetuning for one epoch on $Q$ and pseudo-decompositions $D$ formed via random retrieval, using the final weights to initialize a pretrained encoder-decoder. See Appendix \S\ref{ssec:apdx:Decomposition Training Hyperparameters} for details.

\paragraph{\seqtoseq}
We finetune the pretrained encoder-decoder using maximum likelihood.
We stop training based on validation BLEU between generated decompositions and pseudo-decompositions.

\paragraph{\useqtoseq}
We finetune the pretrained encoder-decoder with back-translation~\cite{sennrich2016improving} and denoising objectives simultaneously, similar to~\citet{lample2019cross} in unsupervised one-to-one translation.\footnote{\url{www.github.com/facebookresearch/XLM}}
For denoising, we produce a noisy input $d'$ by randomly masking, dropping, and locally shuffling tokens in $d \sim D$, and we train a model with parameters $\theta$ to maximize $\log p_{\theta}(d | d')$.
We likewise maximize $\log p_{\theta}(q | q')$ for a noised version $q'$ of $q \sim Q$.
For back-translation, we generate a multi-hop question $\hat{q}$ for a decomposition $d \sim D$, and we maximize $\log p_{\theta}(d | \hat{q})$.
Similarly, we maximize $\log p_{\theta}(q | \hat{d})$ for a model-generated decomposition $\hat{d}$ of $q \sim Q$.
To stop training without supervision, we use a modified version of round-trip BLEU~\cite{lample2018unsupervised} (see Appendix \S\ref{ssec:apdx:unsupervised_stopping_criterion} for details).
We train on \hotpot{} questions $Q$ and their pseudo-decompositions $D$.\footnote{Using the augmented corpora here did not improve QA.}

\subsection{Single-hop Question Answering Model}
We finetune a pretrained model for single-hop QA following prior work from \citet{min2019multi} on \hotpot, as described below.\footnote{Code based on \texttt{transformers}~\cite{wolf2019huggingface}.}

\paragraph{Model Architecture}
Our model takes in a question and several paragraphs to predict the answer.
We compute a separate forward pass on each paragraph (with the question).
For each paragraph, the model learns to predict the answer span if the paragraph contains the answer and to predict ``no answer'' otherwise.
We treat \yes~or \no~predictions as spans within the passage (prepended to each paragraph), as in~\citet{nie2019revealing} on \hotpot{}.
During inference, for the final softmax, we consider all paragraphs as a single chunk.
Similar to~\citet{clark2018simple}, we subtract a paragraph's ``no answer'' logit from the logits of all spans in that paragraph, to reduce or increase span probabilities accordingly.
In other words, we compute the probability $p(s_p)$ of each span $s_p$ in a paragraph $p \in \{ 1, \dots, P \}$ using the predicted span logit $l(s_p)$ and ``no answer'' paragraph logit $n(p)$ with $p(s_p) \propto e^{l(s_p) - n(p)}$.
\robertalarge~\cite{liu2019roberta} is used as our pretrained model.

\paragraph{Training Data and Ensembling}
Similar to~\citet{min2019multi}, we train an ensemble of 2 single-hop QA models on \squad{}~2 and the ``easy'' (single-hop) subset of \hotpot{} (see Appendix \S\ref{sec:apdx:Single-hop QA Model} for training details). We average model logits before predicting the answer.
We use the single-hop QA ensemble as a black-box model once trained, never training the model on multi-hop questions.

\paragraph{Returned Text}
Instead of returning only the predicted sub-answer span to the recomposition model, we return the sentence that contains the predicted sub-answer, which is more informative.

\subsection{Recomposition Model}
Our recomposition model architecture is identical to the single-hop QA model, but the recomposition model also uses sub-questions and sub-answers as input.
We append each (sub-question, sub-answer) pair to the question with separator tokens.
We train one recomposition model on all of \hotpot{}, also including \squad~2 examples used to train the single-hop QA model.
All reported error margins show the mean and std. dev. across 5 recomposition training runs using the same decompositions.

\section{Results on Question Answering}
\begin{table}[t!]
  \centering
  \footnotesize
    \begin{tabular}{ll|ccc}
      Decomp. & Pseudo-  & \multicolumn{3}{c}{\hotpot{} Dev F1} \\
      Method      & Decomps. & Orig & Multi & OOD \\
      \midrule
\xmark & \xmark~(1hop) & 66.7 & 63.7 & 66.5 \\
\xmark & \xmark~(Baseline) & 77.0\textsubscript{$\pm$.2} & 65.2\textsubscript{$\pm$.2} & 67.1\textsubscript{$\pm$.5} \\
\midrule
PseudoD & Random          & 78.4\textsubscript{$\pm$.2} & 70.9\textsubscript{$\pm$.2} & 70.7\textsubscript{$\pm$.4} \\
  & FastText & 78.9\textsubscript{$\pm$.2} & 72.4\textsubscript{$\pm$.1} & 72.0\textsubscript{$\pm$.1} \\
\seqtoseq  & Random          & 77.7\textsubscript{$\pm$.2} & 69.4\textsubscript{$\pm$.3} & 70.0\textsubscript{$\pm$.7} \\
         & FastText        & 78.9\textsubscript{$\pm$.2} & 73.1\textsubscript{$\pm$.2} & 73.0\textsubscript{$\pm$.3} \\
\useqtoseq & Random          & 79.8\textsubscript{$\pm$.1} & 76.0\textsubscript{$\pm$.2} & 76.5\textsubscript{$\pm$.2} \\
        & FastText        & \textbf{80.1}\textsubscript{$\pm$.2} & \textbf{76.2}\textsubscript{$\pm$.1} & \textbf{77.1}\textsubscript{$\pm$.1} \\
\midrule
\multicolumn{2}{l|}{DecompRC*} & 79.8\textsubscript{$\pm$.2} & 76.3\textsubscript{$\pm$.4} & 77.7\textsubscript{$\pm$.2} \\
      \multicolumn{2}{l|}{SAE \cite{tu2020select} $\dagger$}          & 80.2 & 61.1 & 62.6 \\
      \multicolumn{2}{l|}{HGN \cite{fang2019hierarchical} $\dagger$}  & 82.2 & 78.9$\ddagger$ & 76.1$\ddagger$ \\
    \end{tabular}
    \begin{tabular}{l|ccc}
    \toprule
    \toprule
    & Ours & SAE$\dagger$ & HGN$\dagger$ \\
      \midrule
Test (EM/F1) & 66.33/79.34 & 66.92/79.62 & 69.22/82.19 \\
    \end{tabular}
    \caption{Unsupervised decompositions significantly improve F1 on \hotpot{} over the baseline and single-hop QA model used to answer sub-questions (``1hop''). On all dev sets and the test set, we achieve similar F1 to methods that use supporting fact supervision ($\dagger$). (*) We test supervised/heuristic decompositions from~\citet{min2019multi}. ($\ddagger$) Scores are approximate due to mismatched Wikipedia dumps.}
  \label{tab:hotpot_results_main}
\end{table}

We compare variants of our approach that use different learning methods and different pseudo-decomposition training sets.
As a baseline, we compare \roberta{} with decompositions to \roberta{} without decompositions.
We use the best hyperparameters for the baseline to train our \roberta{} models with decompositions (see Appendix \S\ref{ssec:apdx:multi_hop_qa_training_hyperparameters} for hyperparameters).

We report results on 3 dev set versions: (1) the original version,\footnote{Test set is private, so we randomly halve the dev set to form validation/held-out dev sets. Our codebase has our splits.} (2) the multi-hop version from~\citet{jiang2019avoiding} who created some distractor paragraphs adversarially to test multi-hop reasoning, and (3) the out-of-domain (OOD) version from~\citet{min2019multi} who retrieved distractor paragraphs with the same procedure as the original version but excluded the original paragraphs.

\paragraph{Main Results}
Table~\ref{tab:hotpot_results_main} shows how unsupervised decompositions affect QA.
Our \roberta{} baseline does quite well on \hotpot{} (77.0 F1), in line with~\citet{min2019compositional} who achieved strong results using a \bert-based version of the model~\cite{devlin2019bert}.
We achieve large gains over the \roberta{} baseline by simply adding sub-questions and sub-answers to the input.
Using decompositions from \useqtoseq{} trained on FastText pseudo-decompositions, we find a gain of 3.1 F1 on the original dev set, 11 F1 on multi-hop dev, and 10 F1 on OOD dev.
\useqtoseq{} decompositions even match the performance of using supervised and heuristic decompositions from \decomprc{} (i.e., 80.1 vs. 79.8 F1 on the original dev set).

Pseudo-decomposition and \useqtoseq{} training both contribute to decomposition quality.
FastText pseudo-decompositions themselves provide an improvement in QA over the baseline (e.g., 72.0 vs. 67.1 F1 on OOD dev) and over random pseudo-decompositions (70.7 F1), validating our retrieval-based algorithm for creating pseudo-decompositions.
\seqtoseq{} trained on FastText pseudo-decompositions achieves comparable gains to FastText pseudo-decompositions (73.0 F1 on OOD dev), validating the quality of pseudo-decompositions as training data.
As hypothesized, \useqtoseq{} improves over PseudoD and \seqtoseq{} by learning to align hard questions and pseudo-decompositions while ignoring the noisy pairing (77.1 F1 on OOD dev).
\useqtoseq{} is relatively robust to the training data used but still improves further by using FastText vs. Random pseudo-decompositions (77.1 vs. 76.5 F1 on OOD dev).

We submitted the best QA approach based on dev evaluation (using \useqtoseq{} trained on FastText pseudo-decompositions) for hidden test evaluation.
We achieved a test F1 of 79.34 and Exact Match (EM) of 66.33.
Our approach is competitive with state-of-the-art systems SAE~\cite{tu2020select} and HGN~\cite{fang2019hierarchical}, which both (unlike us) learn from strong, supporting-fact supervision about which sentences are relevant to the question.

\begin{table}[!t]
    \begin{minipage}{.5\linewidth}
      \centering
      \footnotesize
    \begin{tabular}{l|cc}
        Q-   & \multicolumn{2}{c}{Using Decomps.} \\
        Type & {\xmark} & {\checkmark} \\
        \midrule
        Bridge     & 80.1\textsubscript{$\pm$.2} & \textbf{81.7}\textsubscript{$\pm$.4} \\
        Comp.      & 73.8\textsubscript{$\pm$.4} & \textbf{80.1}\textsubscript{$\pm$.3} \\
        Inters.  & 79.4\textsubscript{$\pm$.6} & \textbf{82.3}\textsubscript{$\pm$.5} \\
        1-hop & 73.9\textsubscript{$\pm$.6} & \textbf{76.9}\textsubscript{$\pm$.6} \\
    \end{tabular}
    \end{minipage}%
    \begin{minipage}{.5\linewidth}
      \centering
      \footnotesize
    \begin{tabular}{ll|c}
      SQs & SAs & QA F1 \\
      \midrule
      \xmark & \xmark & 77.0\textsubscript{$\pm$.2} \\
      \midrule
      \checkmark & Sent.        & 80.1\textsubscript{$\pm$.2} \\
      \checkmark & Span            & 77.8\textsubscript{$\pm$.3} \\
      \checkmark & Rand.   & 76.9\textsubscript{$\pm$.2} \\
      \checkmark & \xmark          & 76.9\textsubscript{$\pm$.2} \\
      \xmark     & Sent.        & 80.2\textsubscript{$\pm$.1} \\
    \end{tabular}
    \end{minipage}
    \caption{\textbf{Left}: Decompositions improve QA F1 for all 4 \hotpot{} types. \textbf{Right (Ablation)}: QA model F1 when trained with various sub-answers: the sentence of the predicted sub-answer, predicted sub-answer span, or random entity from the context. We also train models with (\checkmark) or without (\xmark) sub-questions/sub-answers.}
    \label{tab:hotpot_results_by_question_type_and_qa_input_ablations}
\end{table}

\subsection{Question Type Breakdown}
\label{ssec:Question Type Breakdown}
To understand where decompositions help, we break down QA accuracy across 4 question types from~\citet{min2019multi}.
``Bridge'' questions ask about an entity not explicitly mentioned (\textit{``When was Erik Watts' father born?''}).
``Intersection'' questions ask to find an entity that satisfies multiple separate conditions (\textit{``Who was on CNBC and Fox News?''}).
``Comparison'' questions ask to compare a property of two entities (\textit{``Which is taller, Momhil Sar or K2?''}).
``Single-hop'' questions are answerable using single-hop shortcuts or single-paragraph reasoning (\textit{``Where is Electric Six from?''}).
We split the original dev set into the 4 types using the supervised type classifier from~\citet{min2019multi}.
Table~\ref{tab:hotpot_results_by_question_type_and_qa_input_ablations} (left) shows F1 scores for \roberta{} with and without decompositions across the 4 types.

\useqtoseq{} decompositions improve QA across all types.
Our single decomposition model does not need to be tailored to the question type, unlike~\citet{min2019multi} who use a different model per question type.
For single-hop questions, our QA approach does not require falling back to a single-hop QA model and instead learns to leverage decompositions in that case also (76.9 vs. 73.9 F1).

\subsection{Answers to Sub-Questions are Crucial}
To measure the usefulness of sub-questions and sub-answers, we train the recomposition model with various, ablated inputs, as shown in Table~\ref{tab:hotpot_results_by_question_type_and_qa_input_ablations} (right).
Sub-answers are crucial to improving QA, as sub-questions with no answers or random answers do not help (76.9 vs. 77.0 F1 for the baseline).
Only when sub-answers are provided do we see improved QA, with or without sub-questions (80.1 and 80.2 F1, respectively).
It is important to provide the sentence containing the predicted answer span instead of the answer span alone (80.1 vs. 77.8 F1, respectively), though the answer span alone still improves over the baseline (77.0 F1).

\begin{figure}[t]
\centering
\includegraphics[width=.92\columnwidth]{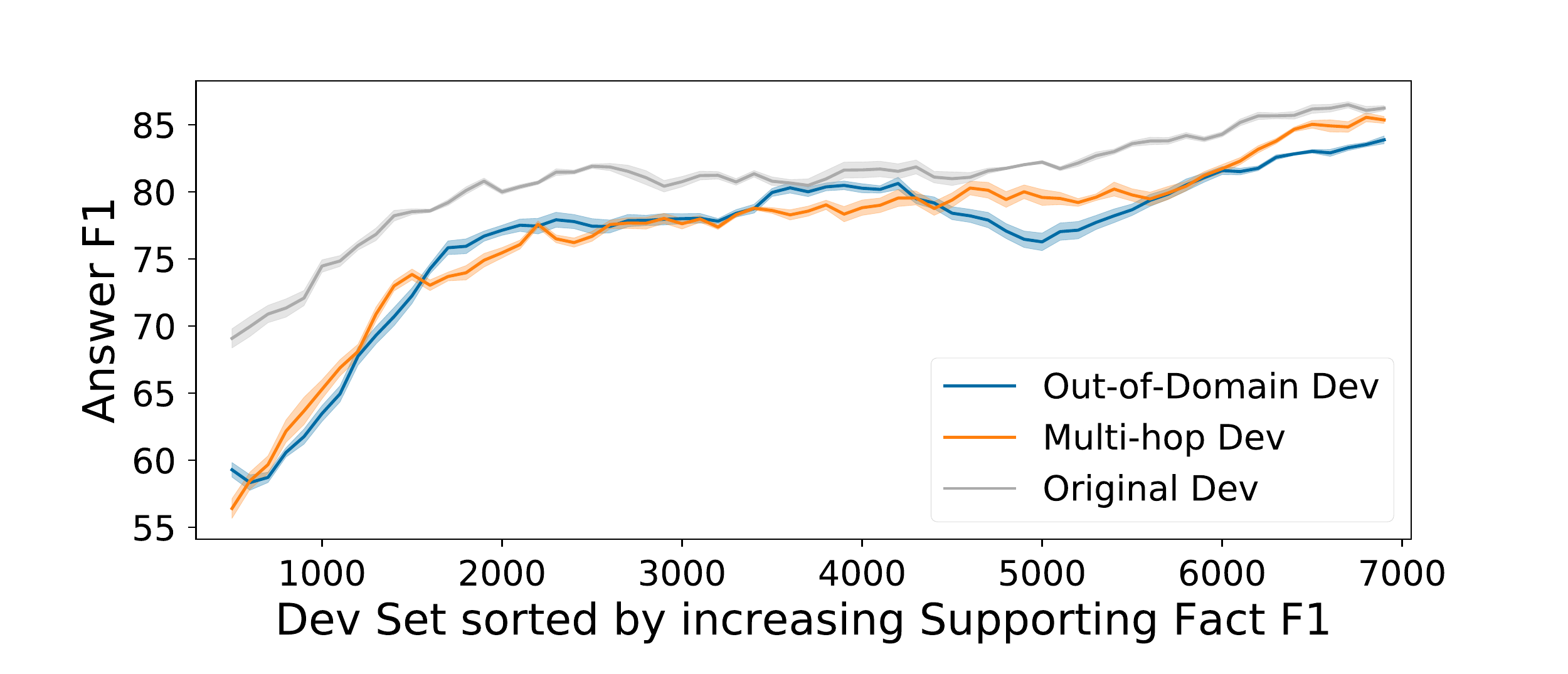}
\caption{Multi-hop QA is better when the single-hop QA model answers with the ground truth ``supporting fact'' sentences. We plot mean and std. over 5 QA runs.}
\label{fig:supporting_fact_f1_vs_answer_f1}
\end{figure}

\subsection{How Do Decompositions Help?}
\label{ssec:how_do_decompositions_help}
Decompositions help by retrieving important supporting evidence to answer questions.
Fig.~\ref{fig:supporting_fact_f1_vs_answer_f1} shows that QA improves when the sub-answer sentences are gold ``supporting facts.''
We retrieve these without relying on strong, supporting fact supervision, unlike many state-of-the-art models~\cite{tu2020select,fang2019hierarchical,nie2019revealing}.\footnote{See Appendix \S\ref{ssec:apdx:Unsupervised Fact Retrieval} for supporting fact scores.}

\subsection{Example Decompositions}
\begin{table}[t!]
    \centering
    \small
    \begin{tabularx}{\columnwidth}{X}
        \toprule
         {\bf Q1}: Who is older, Annie Morton or Terry Richardson?\\
         \hspace{.2cm} {\bf SQ$_1$}: Who is Annie Morton? \\
         \hspace{.25cm} $\llcorner$ Annie Morton (born October 8, 1970) is an\\
         \hspace{.6cm} \underline{American model} born in Pennsylvania. \\
         \hspace{.2cm} {\bf SQ$_2$}: When was Terry Richardson born?\\
         \hspace{.25cm} $\llcorner$ Kenton Terry Richardson (born \underline{26 July 1999)} is an\\
         \hspace{.6cm} English professional footballer who plays as a\\
         \hspace{.6cm} defender for League Two side Hartlepool United.\\
         $\mathbf{\hat{A}}$: Annie Morton \\
        \midrule
         {\bf Q2}: How many copies of Roald Dahl's variation on a popular anecdote sold?\\
         \hspace{.2cm} {\bf SQ$_1$}: How many copies of Roald Dahl's? \\
         \hspace{.25cm} $\llcorner$ His books have sold more than \underline{250 million}\\
         \hspace{.6cm} copies worldwide.\\
         \hspace{.2cm} {\bf SQ$_2$} What is the name of the variation on a\\
         \hspace{.6cm} popular anecdote?\\
         \hspace{.25cm} $\llcorner$ \underline{``Mrs. Bixby and the Colonel's Coat''} is a short story\\
         \hspace{.6cm} by Roald Dahl that first appeared in the 1959 issue of\\
         \hspace{.6cm} Nugget.\\
         $\mathbf{\hat{A}}$: more than 250 million \\
        \midrule
         {\bf Q3}: Are both Coldplay and Pierre Bouvier\\
         \hspace{.55cm} from the same country?\\
         \hspace{.2cm} {\bf SQ$_1$}: Where are Coldplay and Coldplay from? \\
         \hspace{.25cm} $\llcorner$ Coldplay are a \underline{British} rock band formed in 1996 by\\
         \hspace{.6cm} lead vocalist and keyboardist Chris Martin and lead\\
         \hspace{.6cm} guitarist Jonny Buckland at University College\\
         \hspace{.6cm} London (UCL).\\
         \hspace{.2cm} {\bf SQ$_2$}: What country is Pierre Bouvier from?\\
         \hspace{.25cm} $\llcorner$ Pierre Charles Bouvier (born 9 May 1979) is a\\
         \hspace{.6cm} \underline{Canadian} singer, songwriter, musician, composer and\\
         \hspace{.6cm} actor who is best known as the lead singer and\\
         \hspace{.6cm} guitarist of the rock band Simple Plan.\\
         $\mathbf{\hat{A}}$: No \\
        \bottomrule
    \end{tabularx}
    \vskip -0.7em
    \caption{Example sub-questions generated by our model, along with predicted sub-answer sentences (answer span underlined) and final predicted answer.}
    \label{tab:example_decompositions}
\end{table}

To illustrate how decompositions help, Table~\ref{tab:example_decompositions} shows example sub-questions from \useqtoseq{} with predicted sub-answers.
Sub-questions are single-hop questions relevant to the multi-hop question.
The single-hop QA model returns relevant sub-answers, sometimes despite under-specified (\textbf{Q2, SQ$_1$}) or otherwise imperfect sub-questions (\textbf{Q3, SQ$_1$}).
The recomposition model returns an answer consistent with the sub-answers.
Furthermore, the sub-answers used for QA are in natural language, adding a level of interpretability to otherwise black-box, neural QA models.
Decompositions are largely extractive, copying from the multi-hop question rather than hallucinating new entities, which helps generate relevant sub-questions.
Appendix Table~\ref{tab:example_decompositions_other_datasets_zeroshot} shows decompositions from our trained \useqtoseq{} model, without further finetuning, on image-based questions~\citep[CLEVR; ][]{johnson2017clevr}, knowledge-base questions~\citep[ComplexWebQuestions; ][]{talmor2018web}, and even claims in fact verification~\citep[FEVER; ][]{thorne2018fever}, which suggests promising future avenues for our approach in other domains and highlights the general nature of the proposed method.

\section{Analysis}
To better understand our system, we now analyze our pipeline by examining the model for each stage: decomposition, single-hop QA, and recomposition.

\subsection{Unsupervised Decomposition Model}
\paragraph{Intrinsic Evaluation of Decompositions}
\begin{table}[t]
  \centering
  \small
    \begin{tabular}{l|ccccc}
      Decomp. & GPT2 &  \%  Well- & Edit & Length  \\
      Method &NLL & Formed & Dist. & Ratio \\
      \midrule
      \useqtoseq     & 5.56 & 60.9 & 5.96 & 1.08 \\
      DecompRC     & 6.04 & 32.6 & 7.08 & 1.22 \\
    \end{tabular}
    \caption{Analysis of sub-questions produced by our method vs. the supervised+heuristic method of \citet{min2019multi}. Left-to-right: Negative Log-Likelihood according to GPT2 (lower is better), \% classified as Well-Formed, Edit Distance between decomposition and multi-hop question, and token-wise Length Ratio between decomposition and multi-hop question.
    }
  \label{tab:subq_metrics}
\end{table}

We evaluate the quality of decompositions on other metrics aside from downstream QA.
To measure the fluency of decompositions, we compute the likelihood of decompositions using the pretrained GPT-2 language model~\citep{radford2019language}.
We train a \bertbase~classifier on the question-wellformedness dataset of~\citet{faruqui-das-2018-identifying}, and we use the classifier to estimate the proportion of sub-questions that are well-formed.
We measure how abstractive decompositions are by computing (i) the token Levenstein distance between the multi-hop question and its generated decomposition and (ii) the ratio between the length of the decomposition and the length of the multi-hop question.
We compare \useqtoseq{} to \decomprc{}~\cite{min2019multi}, a supervised+heuristic decomposition method.

As shown in Table~\ref{tab:subq_metrics}, \useqtoseq{} decompositions are more natural and well-formed than \decomprc{} decompositions.
As an example, for Table~\ref{tab:example_decompositions} \textbf{Q3}, \decomprc{} produces the sub-questions ``Is Coldplay from which country?'' and ``Is Pierre Bouvier from which country?''
\useqtoseq{} decompositions are also closer in edit distance and length to the multi-hop question, consistent with our observation that our decomposition model is largely extractive.

\paragraph{Quality of Decomposition Model}
\begin{figure}[t]
\centering
\includegraphics[width=\columnwidth]{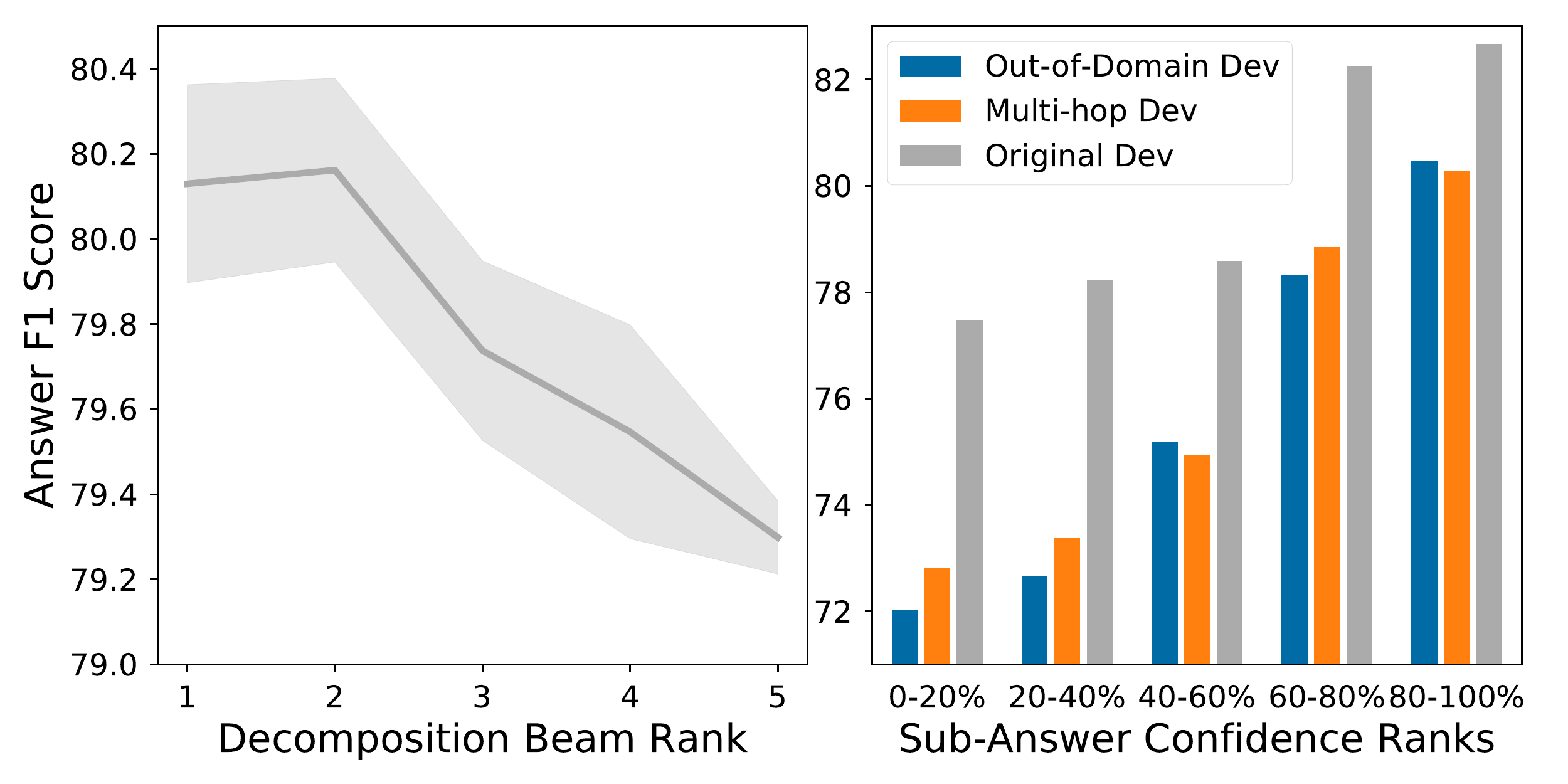}
\caption{\textbf{Left}: We decode decompositions with beam search and use n\textsuperscript{th}-ranked hypothesis as a question decomposition. We plot the F1 of a recomposition model trained to use the n\textsuperscript{th}-ranked decomposition. \textbf{Right}: Multi-hop QA is better when the single-hop QA model places high probability on its sub-answer.}
\label{fig:qa_f1_by_ud_beam_rank_and_answer_confidence}
\end{figure}

A well-trained decomposition model should place higher probability on decompositions that are more helpful for QA.
We generate $N=5$ hypotheses from our best decomposition model using beam search, and we train a recomposition model to use the $n$\textsuperscript{th}-ranked hypothesis as a question decomposition (Figure~\ref{fig:qa_f1_by_ud_beam_rank_and_answer_confidence}, left).
QA accuracy decreases as we use lower probability decompositions, but accuracy remains relatively robust, at most decreasing from 80.1 to 79.3 F1.
The limited drop suggests that decompositions are still useful if they are among the model's top hypotheses, another indication that \useqtoseq{} is trained well for decomposition.

\subsection{Single-hop Question Answering Model}
\paragraph{Sub-Answer Confidence}
Figure \ref{fig:qa_f1_by_ud_beam_rank_and_answer_confidence} (right) shows that the single-hop model's sub-answer confidence correlates with downstream multi-hop QA accuracy on all dev sets.
A low confidence sub-answer may be indicative of (i) an unanswerable or ill-formed sub-question or (ii) a sub-answer that is more likely to be incorrect.
In both cases, the single-hop QA model is less likely to retrieve useful supporting evidence for answering the multi-hop question.

\paragraph{Changing the Single-hop QA Model}
We find that our approach is robust to the single-hop QA model used.
We test the \bertbase{} ensemble from~\citet{min2019multi} as the single-hop QA model.
The model performs much worse compared to our \robertalarge{} single-hop ensemble on \hotpot{} itself (56.3 vs. 66.7 F1).
However, the model results in similar QA when used to answer single-hop sub-questions within our larger system (79.9 vs. 80.1 F1 for our ensemble).

\subsection{Recomposition Model}
\begin{table}[t]
  \centering
  \footnotesize
    \begin{tabular}{l|l}
      Recomposition Model & QA F1 (w/o $\xrightarrow{}$ w/ Decomps.) \\
      \midrule
      \bertbase & 71.8$\pm$.4 $\xrightarrow{}$ 73.0$\pm$.4 \\
      \bertlarge{}       & 76.4$\pm$.2 $\xrightarrow{}$ 79.0$\pm$.1 \\
      \robertalarge{}    & 77.0$\pm$.3 $\xrightarrow{}$ 80.1$\pm$.2 \\
    \end{tabular}
    \caption{
    Better models gain more from decomposition.
    }
  \label{tab:multihop_model_ablation}
\end{table}

\paragraph{Varying the Base Model}
To understand how decompositions impact performance as the recomposition model gets stronger,
we vary the base pretrained model.
Table~\ref{tab:multihop_model_ablation} shows the impact of adding decompositions to \bertbase, \bertlarge, and finally \robertalarge{} (see Appendix \S\ref{ssec:apdx:multi_hop_qa_training_hyperparameters} for hyperparameters).
The gain from using decompositions grows with strength of the recomposition model.
Decompositions improve QA by 1.2 F1 for a \bertbase{} model, by 2.6 F1 for the stronger \bertlarge{} model, and by 3.1 F1 for our best \robertalarge{} model.

\section{Related Work}
Answering complex questions has been a long-standing challenge in natural language processing.
Prior work explored decomposing questions with supervision and heuristic algorithms.
IBM Watson~\cite{ferrucci2010building} decomposes questions into sub-questions in multiple ways or not at all.
\decomprc~\cite{min2019multi} largely frames sub-questions as extractive spans of a question, learning to predict span-based sub-questions via supervised learning on human annotations.
In other cases, \decomprc{} decomposes a multi-hop question using a heuristic algorithm or not at all.
Watson and \decomprc{} use special case handling to decompose different questions, while our algorithm is fully automated and requires little hand-engineering.

More traditional, semantic parsing methods map questions to compositional programs, whose sub-programs can be viewed as question decompositions in a formal language~\cite{talmor2018web,wolfson2020break}.
Examples include classical QA systems like SHRDLU~\cite{winograd1972understanding} and LUNAR~\cite{woods1974the}, as well as neural \seqtoseq{} semantic parsers~\cite{dong2016language} and neural module networks~\cite{andreas2015neural,andreas2016learning}.
Such methods usually require strong, program-level supervision to generate programs, as in visual QA~\cite{johnson2017inferring} and on \hotpot~\cite{jiang2019self-assembling}.
Some models use other forms of strong supervision, e.g., the sentences needed to answer a question, as annotated by \hotpot.
Such an approach is taken by SAE~\cite{tu2020select} and HGN~\cite{fang2019hierarchical}, whose methods may be combined with ours.

Unsupervised decomposition complements strongly and weakly supervised decomposition approaches.
Our unsupervised approach enables methods to leverage millions of otherwise unusable questions, similar to work on unsupervised QA~\cite{lewis2019unsupervised}.
When decomposition examples exist, supervised and unsupervised learning can be used in tandem to learn from both labeled and unlabeled examples.
Such semi-supervised methods outperform supervised learning for tasks like machine translation~\cite{sennrich2016improving}.
Other work on weakly supervised question generation uses a downstream QA model's accuracy as a signal for learning to generate useful questions.
Weakly supervised question generation often uses reinforcement learning~\cite{nogueira2017task,ziyun2019modeling,strub2017end,das2017learning,liang2018memory}, where an unsupervised initialization can greatly mitigate the issues of exploring from scratch~\cite{jaderberg2017reinforcement}.

\section{Conclusion}
We proposed a QA system that answers a question via decomposition, without supervised question decompositions, using three stages: (1) decompose a question into many sub-questions using One-to-N Unsupervised Sequence transduction (\useqtoseq), (2) answer sub-questions with an off-the-shelf QA system, and (3) recompose sub-answers into a final answer.
When evaluated on three \hotpot{} dev sets, our approach significantly improved QA over an equivalent model that did not use decompositions.
Our approach relies only on the final answer as supervision but works as effectively as state-of-the-art methods that rely on much stronger supervision, such as supporting fact labels or example decompositions.
We found that \useqtoseq{} generates fluent sub-questions whose answers often match the gold-annotated, question-relevant text.
Overall, this work opens up exciting avenues for leveraging methods in unsupervised learning and natural language generation to improve the interpretability and generalization of machine learning systems.

\section*{Acknowledgments}
EP's work at NYU is supported by the NSF Graduate Research Fellowship and the Open Philanthropy AI Fellowship.
KC's work at NYU is partly supported by Samsung Advanced Institute of Technology (Next Generation Deep Learning: from pattern recognition to AI) and Samsung Research (Improving Deep Learning using Latent Structure). KC also thanks Naver, eBay, NVIDIA, and NSF Award 1922658 for support. \hotpot{} and \squad{} are licensed under CC BY-SA 4.0. FEVER 1.0 and 2.0 are licensed under CC BY-SA 3.0.
We thank Paul Christiano, Sebastian Riedel, He He, Jonathan Berant, Alexis Conneau, Jiatao Gu, Sewon Min, Yixin Nie, Lajanugen Logeswaran, Adam Fisch, Elman Mansimov, Iacer Calixto, Richard Pang, and our anonymous reviewers for helpful feedback, as well as Yichen Jiang and Peng Qi for help with evaluation.

\bibliography{emnlp2020}

\begin{thebibliography}{52}
\expandafter\ifx\csname natexlab\endcsname\relax\def\natexlab#1{#1}\fi

\bibitem[{Andreas et~al.(2015)Andreas, Rohrbach, Darrell, and
  Klein}]{andreas2015neural}
Jacob Andreas, Marcus Rohrbach, Trevor Darrell, and Dan Klein. 2015.
\newblock \href {https://arxiv.org/abs/1511.02799} {Neural module networks}.
\newblock \emph{CVPR}, pages 39--48.

\bibitem[{Andreas et~al.(2016)Andreas, Rohrbach, Darrell, and
  Klein}]{andreas2016learning}
Jacob Andreas, Marcus Rohrbach, Trevor Darrell, and Dan Klein. 2016.
\newblock \href {https://doi.org/10.18653/v1/N16-1181} {Learning to compose
  neural networks for question answering}.
\newblock In \emph{NAACL}, pages 1545--1554, San Diego, California. Association
  for Computational Linguistics.

\bibitem[{Artetxe et~al.(2018)Artetxe, Labaka, Agirre, and
  Cho}]{artetxe2018unsupervised-neural}
Mikel Artetxe, Gorka Labaka, Eneko Agirre, and Kyunghyun Cho. 2018.
\newblock \href {https://openreview.net/forum?id=Sy2ogebAW} {Unsupervised
  neural machine translation}.
\newblock In \emph{ICLR}.

\bibitem[{Artetxe and Schwenk(2019)}]{artetxe2019margin}
Mikel Artetxe and Holger Schwenk. 2019.
\newblock \href {https://doi.org/10.18653/v1/P19-1309} {Margin-based parallel
  corpus mining with multilingual sentence embeddings}.
\newblock In \emph{ACL}, pages 3197--3203, Florence, Italy. Association for
  Computational Linguistics.

\bibitem[{Bojanowski et~al.(2017)Bojanowski, Grave, Joulin, and
  Mikolov}]{bojanowski-etal-2017-enriching}
Piotr Bojanowski, Edouard Grave, Armand Joulin, and Tomas Mikolov. 2017.
\newblock \href {https://doi.org/10.1162/tacl_a_00051} {Enriching word vectors
  with subword information}.
\newblock \emph{TACL}, 5:135--146.

\bibitem[{Chen et~al.(2017)Chen, Fisch, Weston, and Bordes}]{chen2017reading}
Danqi Chen, Adam Fisch, Jason Weston, and Antoine Bordes. 2017.
\newblock \href {https://doi.org/10.18653/v1/P17-1171} {Reading {W}ikipedia to
  answer open-domain questions}.
\newblock In \emph{ACL}, pages 1870--1879, Vancouver, Canada. Association for
  Computational Linguistics.

\bibitem[{Christiano et~al.(2018)Christiano, Shlegeris, and
  Amodei}]{christiano2018supervising}
Paul~Francis Christiano, Buck Shlegeris, and Dario Amodei. 2018.
\newblock \href {http://arxiv.org/abs/1810.08575} {Supervising strong learners
  by amplifying weak experts}.
\newblock \emph{CoRR}, abs/1810.08575.

\bibitem[{Clark and Gardner(2018)}]{clark2018simple}
Christopher Clark and Matt Gardner. 2018.
\newblock \href {https://doi.org/10.18653/v1/P18-1078} {Simple and effective
  multi-paragraph reading comprehension}.
\newblock In \emph{ACL}, pages 845--855, Melbourne, Australia. Association for
  Computational Linguistics.

\bibitem[{Das et~al.(2017)Das, Kottur, Moura, Lee, and Batra}]{das2017learning}
Abhishek Das, Satwik Kottur, Jos\'e~M.F. Moura, Stefan Lee, and Dhruv Batra.
  2017.
\newblock \href {https://arxiv.org/abs/1511.02799} {Learning cooperative visual
  dialog agents with deep reinforcement learning}.
\newblock In \emph{ICCV}.

\bibitem[{Devlin et~al.(2019)Devlin, Chang, Lee, and
  Toutanova}]{devlin2019bert}
Jacob Devlin, Ming-Wei Chang, Kenton Lee, and Kristina Toutanova. 2019.
\newblock \href {https://doi.org/10.18653/v1/N19-1423} {{BERT}: Pre-training of
  deep bidirectional transformers for language understanding}.
\newblock In \emph{NAACL}, pages 4171--4186.

\bibitem[{Dong and Lapata(2016)}]{dong2016language}
Li~Dong and Mirella Lapata. 2016.
\newblock \href {https://doi.org/10.18653/v1/P16-1004} {Language to logical
  form with neural attention}.
\newblock In \emph{ACL}, pages 33--43, Berlin, Germany. Association for
  Computational Linguistics.

\bibitem[{Fang et~al.(2019)Fang, Sun, Gan, Pillai, Wang, and
  Liu}]{fang2019hierarchical}
Yuwei Fang, Siqi Sun, Zhe Gan, Rohit Pillai, Shuohang Wang, and Jingjing Liu.
  2019.
\newblock \href {http://arxiv.org/abs/1911.03631} {Hierarchical graph network
  for multi-hop question answering}.

\bibitem[{Faruqui and Das(2018)}]{faruqui-das-2018-identifying}
Manaal Faruqui and Dipanjan Das. 2018.
\newblock \href {https://doi.org/10.18653/v1/D18-1091} {Identifying well-formed
  natural language questions}.
\newblock In \emph{EMNLP}, pages 798--803, Brussels, Belgium. Association for
  Computational Linguistics.

\bibitem[{Ferrucci et~al.(2010)Ferrucci, Brown, Chu-Carroll, Fan, Gondek,
  Kalyanpur, Lally, Murdock, Nyberg, Prager, Schlaefer, and
  Welty}]{ferrucci2010building}
David Ferrucci, Eric Brown, Jennifer Chu-Carroll, James Fan, David Gondek,
  Aditya~A. Kalyanpur, Adam Lally, J.~William Murdock, Eric Nyberg, John
  Prager, Nico Schlaefer, and Chris Welty. 2010.
\newblock \href {https://doi.org/10.1609/aimag.v31i3.2303} {Building watson: An
  overview of the deepqa project}.
\newblock \emph{AI Magazine}, 31(3):59--79.

\bibitem[{Guu et~al.(2018)Guu, Hashimoto, Oren, and
  Liang}]{guu-etal-2018-generating}
Kelvin Guu, Tatsunori~B. Hashimoto, Yonatan Oren, and Percy Liang. 2018.
\newblock \href {https://doi.org/10.1162/tacl_a_00030} {Generating sentences by
  editing prototypes}.
\newblock \emph{TACL}, 6:437--450.

\bibitem[{Honnibal and Montani(2017)}]{spacy2}
Matthew Honnibal and Ines Montani. 2017.
\newblock \href {https://github.com/explosion/spaCy} {{spaCy 2}: Natural
  language understanding with {B}loom embeddings, convolutional neural networks
  and incremental parsing}.
\newblock To appear.

\bibitem[{Jaderberg et~al.(2017)Jaderberg, Mnih, Czarnecki, Schaul, Leibo,
  Silver, and Kavukcuoglu}]{jaderberg2017reinforcement}
Max Jaderberg, Volodymyr Mnih, Wojciech~Marian Czarnecki, Tom Schaul, Joel~Z
  Leibo, David Silver, and Koray Kavukcuoglu. 2017.
\newblock \href {https://openreview.net/forum?id=SJ6yPD5xg} {Reinforcement
  learning with unsupervised auxiliary tasks}.
\newblock In \emph{ICLR}.

\bibitem[{Jiang and Bansal(2019{\natexlab{a}})}]{jiang2019avoiding}
Yichen Jiang and Mohit Bansal. 2019{\natexlab{a}}.
\newblock \href {https://doi.org/10.18653/v1/P19-1262} {Avoiding reasoning
  shortcuts: Adversarial evaluation, training, and model development for
  multi-hop {QA}}.
\newblock In \emph{ACL}, pages 2726--2736, Florence, Italy. Association for
  Computational Linguistics.

\bibitem[{Jiang and Bansal(2019{\natexlab{b}})}]{jiang2019self-assembling}
Yichen Jiang and Mohit Bansal. 2019{\natexlab{b}}.
\newblock \href {https://arxiv.org/abs/1909.05803} {Self-assembling modular
  networks for interpretable multi-hop reasoning}.
\newblock In \emph{EMNLP}, Hong Kong, China. Association for Computational
  Linguistics.

\bibitem[{Johnson et~al.(2017{\natexlab{a}})Johnson, Douze, and
  J{\'e}gou}]{JDH17}
Jeff Johnson, Matthijs Douze, and Herv{\'e} J{\'e}gou. 2017{\natexlab{a}}.
\newblock \href {https://arxiv.org/abs/1702.08734} {Billion-scale similarity
  search with gpus}.
\newblock \emph{CoRR}, abs/1702.08734.

\bibitem[{Johnson et~al.(2017{\natexlab{b}})Johnson, Hariharan, van~der Maaten,
  Fei-Fei, Zitnick, and Girshick}]{johnson2017clevr}
Justin Johnson, Bharath Hariharan, Laurens van~der Maaten, Li~Fei-Fei,
  C.~Lawrence Zitnick, and Ross Girshick. 2017{\natexlab{b}}.
\newblock \href {https://arxiv.org/abs/1612.06890} {{CLEVR}: A diagnostic
  dataset for compositional language and elementary visual reasoning}.
\newblock In \emph{CVPR}.

\bibitem[{Johnson et~al.(2017{\natexlab{c}})Johnson, Hariharan, van~der Maaten,
  Hoffman, Fei-Fei, Zitnick, and Girshick}]{johnson2017inferring}
Justin Johnson, Bharath Hariharan, Laurens van~der Maaten, Judy Hoffman,
  Li~Fei-Fei, C~Lawrence Zitnick, and Ross Girshick. 2017{\natexlab{c}}.
\newblock \href {https://arxiv.org/abs/1705.03633} {Inferring and executing
  programs for visual reasoning}.
\newblock In \emph{ICCV}.

\bibitem[{Joulin et~al.(2017)Joulin, Grave, Bojanowski, and
  Mikolov}]{joulin-etal-2017-bag}
Armand Joulin, Edouard Grave, Piotr Bojanowski, and Tomas Mikolov. 2017.
\newblock \href {https://www.aclweb.org/anthology/E17-2068} {Bag of tricks for
  efficient text classification}.
\newblock In \emph{EACL}, pages 427--431, Valencia, Spain. Association for
  Computational Linguistics.

\bibitem[{Lample and Conneau(2019)}]{lample2019cross}
Guillaume Lample and Alexis Conneau. 2019.
\newblock \href {https://arxiv.org/abs/1901.07291} {Cross-lingual language
  model pretraining}.
\newblock In \emph{NeurIPS}.

\bibitem[{Lample et~al.(2018)Lample, Conneau, Denoyer, and
  Ranzato}]{lample2018unsupervised}
Guillaume Lample, Alexis Conneau, Ludovic Denoyer, and Marc'Aurelio Ranzato.
  2018.
\newblock \href {https://openreview.net/forum?id=rkYTTf-AZ} {Unsupervised
  machine translation using monolingual corpora only}.
\newblock In \emph{ICLR}.

\bibitem[{Lewis et~al.(2019)Lewis, Denoyer, and Riedel}]{lewis2019unsupervised}
Patrick Lewis, Ludovic Denoyer, and Sebastian Riedel. 2019.
\newblock \href {https://doi.org/10.18653/v1/P19-1484} {Unsupervised question
  answering by cloze translation}.
\newblock In \emph{ACL}, pages 4896--4910, Florence, Italy. Association for
  Computational Linguistics.

\bibitem[{Liang et~al.(2018)Liang, Norouzi, Berant, Le, and
  Lao}]{liang2018memory}
Chen Liang, Mohammad Norouzi, Jonathan Berant, Quoc~V Le, and Ni~Lao. 2018.
\newblock \href {https://arxiv.org/abs/1807.02322} {Memory augmented policy
  optimization for program synthesis and semantic parsing}.
\newblock In S.~Bengio, H.~Wallach, H.~Larochelle, K.~Grauman, N.~Cesa-Bianchi,
  and R.~Garnett, editors, \emph{NeurIPS}, pages 9994--10006. Curran
  Associates, Inc.

\bibitem[{Liu et~al.(2019)Liu, Ott, Goyal, Du, Joshi, Chen, Levy, Lewis,
  Zettlemoyer, and Stoyanov}]{liu2019roberta}
Yinhan Liu, Myle Ott, Naman Goyal, Jingfei Du, Mandar Joshi, Danqi Chen, Omer
  Levy, Mike Lewis, Luke Zettlemoyer, and Veselin Stoyanov. 2019.
\newblock \href {https://arxiv.org/abs/1907.11692} {{RoBERTa}: A robustly
  optimized bert pretraining approach}.
\newblock \emph{CoRR}, abs/1907.11692.

\bibitem[{Micikevicius et~al.(2018)Micikevicius, Narang, Alben, Diamos, Elsen,
  Garcia, Ginsburg, Houston, Kuchaiev, Venkatesh, and
  Wu}]{micikevicius2018mixed}
Paulius Micikevicius, Sharan Narang, Jonah Alben, Gregory Diamos, Erich Elsen,
  David Garcia, Boris Ginsburg, Michael Houston, Oleksii Kuchaiev, Ganesh
  Venkatesh, and Hao Wu. 2018.
\newblock \href {https://openreview.net/forum?id=r1gs9JgRZ} {Mixed precision
  training}.
\newblock In \emph{ICLR}.

\bibitem[{Min et~al.(2019{\natexlab{a}})Min, Wallace, Singh, Gardner,
  Hajishirzi, and Zettlemoyer}]{min2019compositional}
Sewon Min, Eric Wallace, Sameer Singh, Matt Gardner, Hannaneh Hajishirzi, and
  Luke Zettlemoyer. 2019{\natexlab{a}}.
\newblock \href {https://doi.org/10.18653/v1/P19-1416} {Compositional questions
  do not necessitate multi-hop reasoning}.
\newblock In \emph{ACL}, pages 4249--4257, Florence, Italy. Association for
  Computational Linguistics.

\bibitem[{Min et~al.(2019{\natexlab{b}})Min, Zhong, Zettlemoyer, and
  Hajishirzi}]{min2019multi}
Sewon Min, Victor Zhong, Luke Zettlemoyer, and Hannaneh Hajishirzi.
  2019{\natexlab{b}}.
\newblock \href {https://doi.org/10.18653/v1/P19-1613} {Multi-hop reading
  comprehension through question decomposition and rescoring}.
\newblock In \emph{ACL}, pages 6097--6109, Florence, Italy. Association for
  Computational Linguistics.

\bibitem[{Nie et~al.(2019)Nie, Wang, and Bansal}]{nie2019revealing}
Yixin Nie, Songhe Wang, and Mohit Bansal. 2019.
\newblock \href {https://www.aclweb.org/anthology/D19-1258/} {Revealing the
  importance of semantic retrieval for machine reading at scale}.
\newblock In \emph{EMNLP}.

\bibitem[{Nogueira and Cho(2017)}]{nogueira2017task}
Rodrigo Nogueira and Kyunghyun Cho. 2017.
\newblock \href {https://doi.org/10.18653/v1/D17-1061} {Task-oriented query
  reformulation with reinforcement learning}.
\newblock In \emph{EMNLP}, pages 574--583, Copenhagen, Denmark. Association for
  Computational Linguistics.

\bibitem[{Petrochuk and Zettlemoyer(2018)}]{petrochuk2018simplequestions}
Michael Petrochuk and Luke Zettlemoyer. 2018.
\newblock \href {https://doi.org/10.18653/v1/D18-1051} {{S}imple{Q}uestions
  nearly solved: A new upperbound and baseline approach}.
\newblock In \emph{EMNLP}, pages 554--558, Brussels, Belgium. Association for
  Computational Linguistics.

\bibitem[{Radford et~al.(2019)Radford, Wu, Child, Luan, Amodei, and
  Sutskever}]{radford2019language}
Alec Radford, Jeff Wu, Rewon Child, David Luan, Dario Amodei, and Ilya
  Sutskever. 2019.
\newblock \href
  {https://d4mucfpksywv.cloudfront.net/better-language-models/language_models_are_unsupervised_multitask_learners.pdf}
  {Language models are unsupervised multitask learners}.

\bibitem[{Sennrich et~al.(2016)Sennrich, Haddow, and
  Birch}]{sennrich2016improving}
Rico Sennrich, Barry Haddow, and Alexandra Birch. 2016.
\newblock \href {https://doi.org/10.18653/v1/P16-1009} {Improving neural
  machine translation models with monolingual data}.
\newblock In \emph{ACL}, pages 86--96, Berlin, Germany. Association for
  Computational Linguistics.

\bibitem[{Strub et~al.(2017)Strub, de~Vries, Mary, Piot, Courville, and
  Pietquin}]{strub2017end}
Florian Strub, Harm de~Vries, Jérémie Mary, Bilal Piot, Aaron Courville, and
  Olivier Pietquin. 2017.
\newblock \href {https://doi.org/10.24963/ijcai.2017/385} {End-to-end
  optimization of goal-driven and visually grounded dialogue systems}.
\newblock In \emph{IJCAI}, pages 2765--2771.

\bibitem[{Talmor and Berant(2018)}]{talmor2018web}
Alon Talmor and Jonathan Berant. 2018.
\newblock \href {https://doi.org/10.18653/v1/N18-1059} {The web as a
  knowledge-base for answering complex questions}.
\newblock In \emph{NAACL}, pages 641--651, New Orleans, Louisiana. Association
  for Computational Linguistics.

\bibitem[{Thorne et~al.(2018)Thorne, Vlachos, Christodoulopoulos, and
  Mittal}]{thorne2018fever}
James Thorne, Andreas Vlachos, Christos Christodoulopoulos, and Arpit Mittal.
  2018.
\newblock \href {http://arxiv.org/abs/1803.05355} {{FEVER:} a large-scale
  dataset for fact extraction and verification}.
\newblock \emph{CoRR}, abs/1803.05355.

\bibitem[{Thorne et~al.(2019)Thorne, Vlachos, Cocarascu, Christodoulopoulos,
  and Mittal}]{thorne2019fever2}
James Thorne, Andreas Vlachos, Oana Cocarascu, Christos Christodoulopoulos, and
  Arpit Mittal. 2019.
\newblock \href {https://doi.org/10.18653/v1/D19-6601} {The {FEVER}2.0 shared
  task}.
\newblock In \emph{Proceedings of the Second Workshop on Fact Extraction and
  VERification (FEVER)}, pages 1--6, Hong Kong, China. Association for
  Computational Linguistics.

\bibitem[{Tu et~al.(2020)Tu, Huang, Wang, Huang, He, and Zhou}]{tu2020select}
Ming Tu, Kevin Huang, Guangtao Wang, Jing Huang, Xiaodong He, and Bowen Zhou.
  2020.
\newblock \href {https://arxiv.org/abs/1911.00484} {Select, answer and explain:
  Interpretable multi-hop reading comprehension over multiple documents}.
\newblock In \emph{{AAAI}}.

\bibitem[{Vaswani et~al.(2017)Vaswani, Shazeer, Parmar, Uszkoreit, Jones,
  Gomez, Kaiser, and Polosukhin}]{vaswani2018attention}
Ashish Vaswani, Noam Shazeer, Niki Parmar, Jakob Uszkoreit, Llion Jones,
  Aidan~N Gomez, \L~ukasz Kaiser, and Illia Polosukhin. 2017.
\newblock \href
  {http://papers.nips.cc/paper/7181-attention-is-all-you-need.pdf} {Attention
  is all you need}.
\newblock In I.~Guyon, U.~V. Luxburg, S.~Bengio, H.~Wallach, R.~Fergus,
  S.~Vishwanathan, and R.~Garnett, editors, \emph{NeurIPS}, pages 5998--6008.
  Curran Associates, Inc.

\bibitem[{Wang and Lake(2019)}]{ziyun2019modeling}
Ziyun Wang and Brenden~M. Lake. 2019.
\newblock \href {http://arxiv.org/abs/1907.09899} {Modeling question asking
  using neural program generation}.
\newblock \emph{CoRR}, abs/1907.09899.

\bibitem[{Winograd(1972)}]{winograd1972understanding}
Terry Winograd. 1972.
\newblock \href
  {https://www.sciencedirect.com/science/article/pii/0010028572900023}
  {\emph{Understanding Natural Language}}.
\newblock Academic Press, Inc., USA.

\bibitem[{Winograd(1991)}]{winograd1991thinking}
Terry Winograd. 1991.
\newblock \href
  {http://hci.stanford.edu/~winograd/papers/thinking-machines.html}
  {\emph{Thinking Machines: Can There Be? Are We?}}
\newblock University of California Press, Berkeley.

\bibitem[{Wolf et~al.(2019)Wolf, Debut, Sanh, Chaumond, Delangue, Moi, Cistac,
  Rault, Louf, Funtowicz, and Brew}]{wolf2019huggingface}
Thomas Wolf, Lysandre Debut, Victor Sanh, Julien Chaumond, Clement Delangue,
  Anthony Moi, Pierric Cistac, Tim Rault, R'emi Louf, Morgan Funtowicz, and
  Jamie Brew. 2019.
\newblock \href {https://arxiv.org/abs/1910.03771} {Huggingface's transformers:
  State-of-the-art natural language processing}.
\newblock \emph{CoRR}, abs/1910.03771.

\bibitem[{Wolfson et~al.(2020)Wolfson, Geva, Gupta, Gardner, Goldberg, Deutch,
  and Berant}]{wolfson2020break}
Tomer Wolfson, Mor Geva, Ankit Gupta, Matt Gardner, Yoav Goldberg, Daniel
  Deutch, and Jonathan Berant. 2020.
\newblock \href {https://arxiv.org/abs/2001.11770} {Break it down: A question
  understanding benchmark}.
\newblock \emph{TACL}.

\bibitem[{Woods et~al.(1974)Woods, Kaplan, and Nash-Webber}]{woods1974the}
W.~Woods, R.~Kaplan, and B.~Nash-Webber. 1974.
\newblock \href {https://dl.acm.org/doi/10.1145/1499586.1499695} {The lunar
  sciences natural language information system}.
\newblock Final Report 2378, Bolt, Beranek and Newman, Inc., Cambridge, MA.

\bibitem[{Xu and Koehn(2017)}]{xu2017zipporah}
Hainan Xu and Philipp Koehn. 2017.
\newblock \href {https://doi.org/10.18653/v1/D17-1319} {{Z}ipporah: a fast and
  scalable data cleaning system for noisy web-crawled parallel corpora}.
\newblock In \emph{EMNLP}, pages 2945--2950, Copenhagen, Denmark. Association
  for Computational Linguistics.

\bibitem[{Yang et~al.(2018)Yang, Qi, Zhang, Bengio, Cohen, Salakhutdinov, and
  Manning}]{yang2018hotpotqa}
Zhilin Yang, Peng Qi, Saizheng Zhang, Yoshua Bengio, William Cohen, Ruslan
  Salakhutdinov, and Christopher~D. Manning. 2018.
\newblock \href {https://doi.org/10.18653/v1/D18-1259} {{H}otpot{QA}: A dataset
  for diverse, explainable multi-hop question answering}.
\newblock In \emph{EMNLP}, pages 2369--2380, Brussels, Belgium. Association for
  Computational Linguistics.

\bibitem[{Zhou et~al.(2015)Zhou, He, Zhao, and Hu}]{zhou2015learning}
Guangyou Zhou, Tingting He, Jun Zhao, and Po~Hu. 2015.
\newblock \href {https://doi.org/10.3115/v1/P15-1025} {Learning continuous word
  embedding with metadata for question retrieval in community question
  answering}.
\newblock In \emph{ACL}, pages 250--259, Beijing, China. Association for
  Computational Linguistics.

\bibitem[{Zhu et~al.(2015)Zhu, Kiros, Zemel, Salakhutdinov, Urtasun, Torralba,
  and Fidler}]{zhu2015aligning}
Yukun Zhu, Ryan Kiros, Rich Zemel, Ruslan Salakhutdinov, Raquel Urtasun,
  Antonio Torralba, and Sanja Fidler. 2015.
\newblock \href {https://doi.org/10.1109/ICCV.2015.11} {Aligning books and
  movies: Towards story-like visual explanations by watching movies and reading
  books}.
\newblock In \emph{ICCV}, page 19–27, USA. IEEE Computer Society.

\end{thebibliography}
\bibliographystyle{acl_natbib}

\clearpage
\appendix



\section{Pseudo-Decompositions}

Tables~\ref{tab:appendix_decomp_examples_1}-\ref{tab:appendix_decomp_examples_3} show examples of pseudo-decompositions and learned decompositions from various models.

\subsection{Variable-Length Pseudo-Decompositions}
\label{ssec:apdx:generalized_pseudo_decomposition}
A general algorithm for creating pseudo-decompositions should find a suitable number of sub-questions $N$ for each question.
To this end, we compare the objective in Eq.~\ref{eqn:generalized_similarity_retrieval_eqn} for creating pseudo-decompositions with an alternate objective based on Euclidean distance.
This alternate objective has the advantage that the regularization term that encourages sub-question diversity grows more slowly $N$, disencouraging larger $N$ less:
\begin{equation}
\label{eqn:generalized_similarity_retrieval_eqn_euclid}
  d'^* = \argmin_{d' \subset S} \left|\left| \mathbf{v}_{q} - \sum_{s \in d'}\mathbf{v}_{s} \right|\right|_2
\end{equation}

We create pseudo-decompositions in an similar way as with Eq.~\ref{eqn:generalized_similarity_retrieval_eqn}, first finding a set of candidate sub-questions $S' \subset S$ with high cosine similarity to $\mathbf{v}_q$.
Then, we perform beam search to sequentially choose sub-questions up to a maximum of $N$ sub-questions.

We test pseudo-decomposition objectives by creating synthetic, compositional questions by combining 2-3 single-hop questions with ``and.''
Then, we measure rank of the correct decomposition (a concatenation of the single-hop questions), according to each objective.
For $N=2$, both objectives perform well.
For $N=3$, Eq. \ref{eqn:generalized_similarity_retrieval_eqn_euclid} achieves a mean reciprocal rank of 30\%, while Eq. \ref{eqn:generalized_similarity_retrieval_eqn} gets $\sim$0\%.
In practice, few questions appear to require $N>2$ on \hotpot, as we find similar QA accuracy with Eq. \ref{eqn:generalized_similarity_retrieval_eqn} (which consistently uses $N=2$ sub-questions) and Eq. \ref{eqn:generalized_similarity_retrieval_eqn_euclid} (which mostly uses $N=2$ but sometimes uses $N=3$).
For example, with Eq. \ref{eqn:generalized_similarity_retrieval_eqn} vs. Eq. \ref{eqn:generalized_similarity_retrieval_eqn_euclid}, we find 79.9 vs. 79.4 dev F1 when using the \bertbase{} ensemble from~\citet{min2019multi} to answer sub-questions.
Thus, we use Eq. \ref{eqn:generalized_similarity_retrieval_eqn} in our main experiments, as it is simpler and faster to compute.
Table~\ref{tab:appendix_decomp_examples_1} contains an example where the variable-length decomposition method discussed above (Eq. \ref{eqn:generalized_similarity_retrieval_eqn_euclid}) generates three sub-questions while other methods produce two.


\subsection{Impact of Question Corpus Size}
In addition to our previous results on FastText vs. Random pseudo-decompositions, we found it important to use a large question corpus to create pseudo-decompositions.
QA F1 increased from 79.2 to 80.1 when we trained decomposition models on pseudo-decompositions comprised of questions retrieved from Common Crawl ($>$10M questions) rather than only \squad{} 2 ($\sim$130K questions), using an appropriately larger beam size for pseudo-decomposition (100 $\rightarrow$ 1000).

\subsection{Question Mining Details}
\label{ssec:apdx:Question Mining Details}
We train a 4-way FastText, bag-of-words classifier to classifier between (1) \hotpot{} ``Bridge''/``Intersection'' questions (See \S\ref{ssec:Question Type Breakdown} for definitions), (2) \hotpot{} ``Comparison'' questions (See \S\ref{ssec:Question Type Breakdown} for definition), (3) SQuAD 2.0 questions, (4) and Common Crawl questions.
We randomly sample 15K examples from each of the above four groups of questions to form our training data.
The trained classifier performs well, achieving 95.5\% accuracy for \hotpot{} vs. SQuAD question classification on held-out questions.
Questions in Common Crawl that were classified as from \hotpot{} by the classifier often had more words, conjunctions (``or,'' ``and''), and comparison words (``older,'' ``earlier''), and were generally complex questions.

\subsection{Pseudo-Decomposition Retrieval Method}
\label{ssec:apdx:pseudo_decomposition_retrieval_method}
\begin{table}[t!]
  \centering
  \footnotesize
    \begin{tabular}{ll|ccc}
      Decomp. & Pseudo-  & \multicolumn{3}{c}{\hotpot{} F1} \\
      Method      & Decomps. & Dev & Advers. & OOD \\
      \midrule
\xmark & \xmark~(1hop) & 66.7 & 63.7 & 66.5 \\
\xmark & \xmark~(Baseline) & 77.0$\pm$\textsubscript{.2} & 65.2$\pm$\textsubscript{.2} & 67.1$\pm$\textsubscript{.5} \\
\midrule
PseudoD & Random          & 78.4$\pm$\textsubscript{.2} & 70.9$\pm$\textsubscript{.2} & 70.7$\pm$\textsubscript{.4} \\
      &\bert{}   & 78.9$\pm$\textsubscript{.4} & 71.5$\pm$\textsubscript{.3} & 71.5$\pm$\textsubscript{.2} \\
      & TFIDF    & 79.2$\pm$\textsubscript{.3} & 72.2$\pm$\textsubscript{.3} & 72.0$\pm$\textsubscript{.5} \\
      & FastText & 78.9$\pm$\textsubscript{.2} & 72.4$\pm$\textsubscript{.1} & 72.0$\pm$\textsubscript{.1} \\
\seqtoseq & Random   & 77.7$\pm$\textsubscript{.2} & 69.4$\pm$\textsubscript{.3} & 70.0$\pm$\textsubscript{.7} \\
      & \bert{}  & 79.1$\pm$\textsubscript{.3} & 72.6$\pm$\textsubscript{.3} & 73.1$\pm$\textsubscript{.3} \\
      & TFIDF    & 79.2$\pm$\textsubscript{.1} & 73.0$\pm$\textsubscript{.3} & 72.9$\pm$\textsubscript{.3} \\
      & FastText & 78.9$\pm$\textsubscript{.2} & 73.1$\pm$\textsubscript{.2} & 73.0$\pm$\textsubscript{.3} \\
\cseqtoseq & Random   & 79.4$\pm$\textsubscript{.2} & 75.1$\pm$\textsubscript{.2} & 75.2$\pm$\textsubscript{.4} \\
      & \bert{}  & 78.9$\pm$\textsubscript{.2} & 74.9$\pm$\textsubscript{.1} & 75.2$\pm$\textsubscript{.2} \\
      & TFIDF    & 78.6$\pm$\textsubscript{.3} & 72.4$\pm$\textsubscript{.4} & 72.8$\pm$\textsubscript{.2} \\
      & FastText & 79.9$\pm$\textsubscript{.2} & 76.0$\pm$\textsubscript{.1} & 76.9$\pm$\textsubscript{.1} \\
\useqtoseq & Random   & 79.8$\pm$\textsubscript{.1} & 76.0$\pm$\textsubscript{.2} & 76.5$\pm$\textsubscript{.2} \\
      & \bert{}  & 79.8$\pm$\textsubscript{.3} & \textbf{76.2}$\pm$\textsubscript{.3} & 76.7$\pm$\textsubscript{.3} \\
      & TFIDF    & 79.6$\pm$\textsubscript{.2} & 75.5$\pm$\textsubscript{.2} & 76.0$\pm$\textsubscript{.2} \\
      & FastText & \textbf{80.1}$\pm$\textsubscript{.2} & \textbf{76.2}$\pm$\textsubscript{.1} & \textbf{77.1}$\pm$\textsubscript{.1} \\
\midrule
      \multicolumn{2}{l|}{DecompRC} & 79.8$\pm$\textsubscript{.2} & 76.3$\pm$\textsubscript{.4} & 77.7$\pm$\textsubscript{.2} \\
      \multicolumn{2}{l|}{SAE \cite{tu2020select}}       & 80.2 & 61.1 & 62.6 \\
      \multicolumn{2}{l|}{HGN \cite{fang2019hierarchical}} & 82.2 & 78.9 & 76.1 \\
    \end{tabular}
    \caption{QA F1 scores for all combinations of learning methods and pseudo-decomposition retrieval methods that we tried.
    }
  \label{tab:hotpot_results_full_simple}
\end{table}

Table~\ref{tab:hotpot_results_full_simple} shows QA results with pseudo-decompositions retrieved using sum-bag-of-word representations from FastText, TFIDF, \bertlarge{} first layer hidden states. 
We also vary the learning method and include results Curriculum \useqtoseq{} (\cseqtoseq), where we initialize the \useqtoseq{} approach with the \seqtoseq{} model trained on the same data.

\section{Unsupervised Decomposition Model}
\begin{figure}[t!]
\centering
\includegraphics[width=\columnwidth]{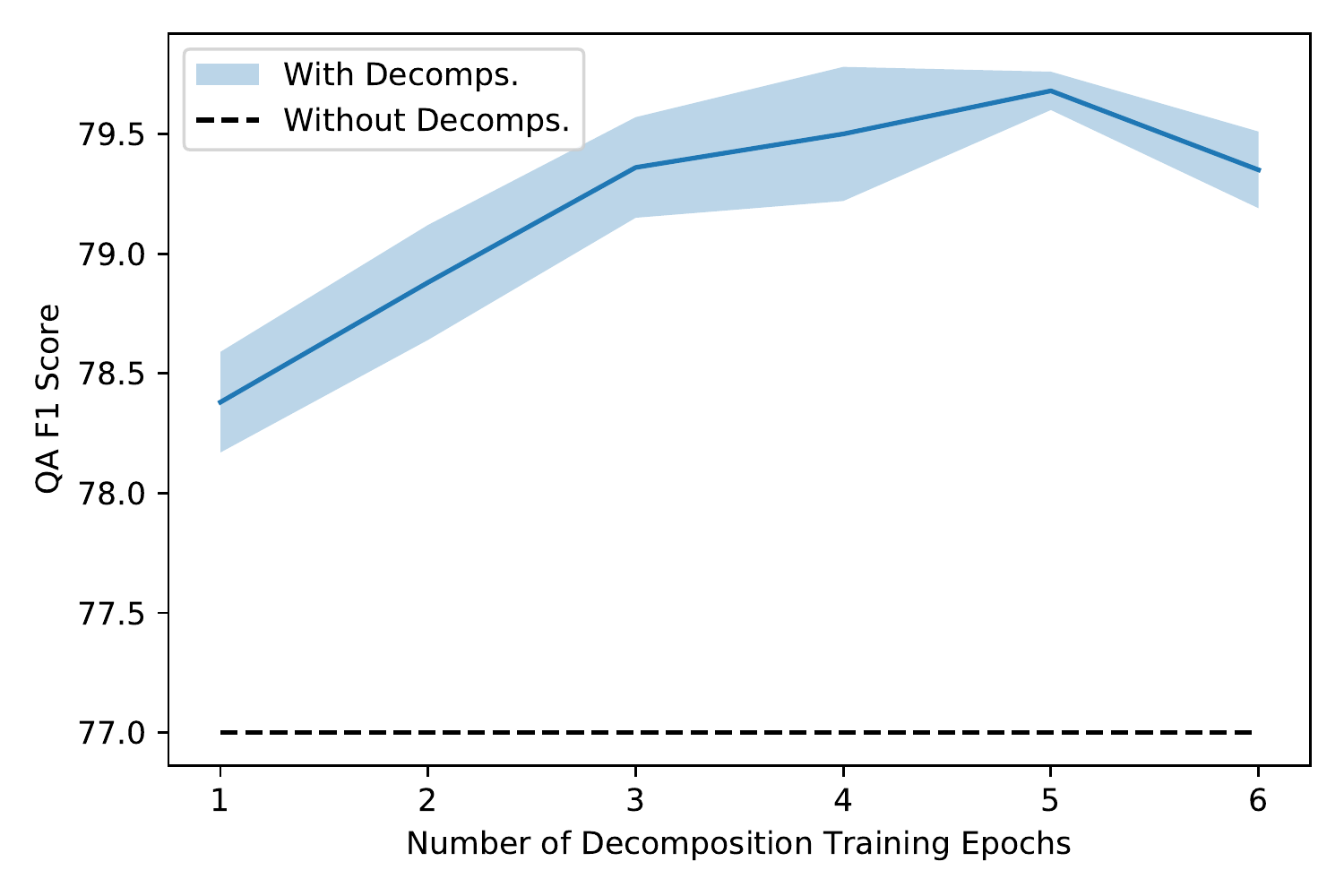}
\caption{How multi-hop QA accuracy varies over the course of decomposition model training, for one training run of \useqtoseq{} on FastText pseudo-decompositions. Our unsupervised stopping criterion selects the epoch 3 checkpoint, which performs roughly as well as the best checkpoint (epoch 5).}
\label{fig:QA_F1_across_UD_Training}
\end{figure}

\subsection{Training Procedure}
\label{ssec:apdx:unsupervised_stopping_criterion}
\paragraph{Unsupervised Stopping Criterion}
To stop \useqtoseq{} training, we use an unsupervised stopping criterion to avoid relying on a supervised validation set of decompositions.
We generate a decomposition $\hat{d}$ for a multi-hop question $q$, and we measure BLEU between $q$ and the model-generated question $\hat{q}$ for $\hat{d}$, similar to round-trip BLEU in unsupervised one-to-one translation~\cite{lample2018unsupervised}.
We scale round-trip BLEU score by the fraction of ``good'' decompositions, where a good decomposition has (1) two sub-questions (question marks), (2) no sub-question which contains all words in the multi-hop question, and (3) no sub-question longer than the multi-hop question.
We chose these criteria to detect a failure mode; without scaling, decomposition models can achieve perfect round-trip BLEU by copying the multi-hop question as the decomposition.
We measure scaled BLEU across multi-hop questions in \hotpot{} dev, and we stop training when the metric does not increase for 3 consecutive epochs.

It is possible to stop training the decomposition model based on downstream QA accuracy.
However, training a QA model on each decomposition model checkpoint (1) is computationally expensive and (2) ties decompositions to a specific, downstream QA model.
In Figure~\ref{fig:QA_F1_across_UD_Training}, we show downstream QA results across various \useqtoseq{} checkpoints when using the \bertbase{} single-hop QA ensemble from~\citet{min2019multi}. The unsupervised stopping criterion does not significantly hurt downstream QA compared to using a weakly-supervised stopping criterion based on multi-hop QA accuracy.

\subsection{Training Hyperparameters}
\label{ssec:apdx:Decomposition Training Hyperparameters}

\paragraph{MLM Pretraining}
We warm-start our pretraining with the 340M parameter, pretrained, English Masked Language Model (MLM) from~\citet{lample2019cross}, a 12-block \textit{encoder-only} transformer~\citep{vaswani2018attention} trained on Toronto Books Corpus~\citep{zhu2015aligning} and Wikipedia.
We pretrain our encoder for 26 hours (one full epoch on $Q$) with 8 DGX-1 machines, each with 8, 32GB NVIDIA V100 GPUs interconnected by Infiniband.
We use the largest possible batch size ($1536$), and we choose the best learning rate ($3 \times 10^{-5}$) based on training loss after a small number of iterations.
We chose a maximum sequence length of $128$.
Other hyperparameters are identical to those from~\citet{lample2019cross} used in unsupervised one-to-one translation.
To initialize a pretrained \textit{encoder-decoder} from the encoder-only MLM, we initialize a 6-block encoder with the first 6 MLM blocks, and we initialize a 6-block decoder with the last 6 MLM blocks, randomly initializing the remaining weights as in~\citet{lample2019cross}.

\paragraph{\useqtoseq}
We train each decomposition model with distributed training over 8, 32GB NVIDIA V100 GPUs, lasting roughly 8 hours.
We chose the largest batch size that fit in GPU memory ($256$) and then the largest learning rate that resulted in stable learning early in training ($3 \times 10^{-5}$).
Other hyperparameters are the same as~\citet{lample2019cross}.

\paragraph{\seqtoseq}
We again train each decomposition model with distributed training over 8, 32GB NVIDIA V100 GPUs, lasting roughly 8 hours.
We use a large batch size ($1024$) and chose the largest learning rate which resulted in stable training across the various pseudo-decomposition training corpora from Appendix \S\ref{ssec:apdx:pseudo_decomposition_retrieval_method} ($1 \times 10^{-4}$).
We keep other training settings and hyperparameters the same as for \useqtoseq.

\subsection{Unsupervised Fact Retrieval}
\label{ssec:apdx:Unsupervised Fact Retrieval}
Our unsupervised supporting fact retrieval (described in \S\ref{ssec:how_do_decompositions_help}) achieves 15.7 EM and 55.2 F1 for retrieving the gold supporting facts (sentences) needed to answer \hotpot{} questions.
To our knowledge, there is no prior work on unsupervised fact retrieval on \hotpot{} to compare against, but our performance approaches early, supervised fact-retrieval methods on \hotpot{} from~\citet{yang2018hotpotqa} which achieve 59.0 F1.

\subsection{Decomposing Questions in Other Tasks}
\begin{table*}[t]
  \centering
  \small
    \begin{tabular}{l|l}
      \textbf{Dataset} & \textbf{Question and \useqtoseq{} Decomposition} \\
      \midrule
         \textbf{FEVER 1.0} & {\bf Q1}: The highest point of the Hindu Kush is Everest.\\
         & \hspace{.3cm} {\bf SQ$_1$}: The highest point of the Hindu Kush?\\
         & \hspace{.3cm} {\bf SQ$_2$}: Where is Everest?\\
         & {\bf Q2}: John Dolmayan was born on July 15, 1873.\\
         & \hspace{.3cm} {\bf SQ$_1$}: When was John Dolmayan born?\\
         & \hspace{.3cm} {\bf SQ$_2$}: Who was born on July 15, 1873.?\\
         & {\bf Q3}: Colin Kaepernick became a starter during the 49ers 63rd season in the Republican party.\\
         & \hspace{.3cm} {\bf SQ$_1$}: When did Colin Kaepernick become a starter?\\
         & \hspace{.3cm} {\bf SQ$_2$}: The 49ers 63rd season in the Republican party.?\\
         & {\bf Q4}: Buffy Summers has been written by Sarah Michelle Gellar.\\
         & \hspace{.3cm} {\bf SQ$_1$}: When has Buffy Summers been written?\\
         & \hspace{.3cm} {\bf SQ$_2$}: Who was Sarah Michelle Gellar.?\\
         \midrule
         \textbf{FEVER 2.0} & {\bf Q1}: Brad Wilk co-founded Rage with Tom Morello and Zack de la Rocha before 1940.\\
         & \hspace{.3cm} {\bf SQ$_1$}: When did Brad Wilk co-founded Rage with Tom Morello?\\
         & \hspace{.3cm} {\bf SQ$_2$}: Who was Zack de la Rocha before 1940?\\
         & {\bf Q2}: David Spade starred in a 2015 American comedy film directed by Fred Wolf\\
         & \hspace{.3cm} {\bf SQ$_1$}: When was David Spade born?\\
         & \hspace{.3cm} {\bf SQ$_2$}: Who directed the 2015 American comedy film?\\
         & {\bf Q3}: Java is in Indonesia and was formed by volcanic eruptions Pleistocene Era.\\
         & \hspace{.3cm} {\bf SQ$_1$}: Where is Java in Indonesia?\\
         & \hspace{.3cm} {\bf SQ$_2$}: When were the last volcanic eruptions of Pleistocene Era.\\
         & {\bf Q4}: Henry Cavill played a fictional character, a superhero appearing\\
         & \hspace{.8cm} in American comic books published by DC Comics.\\
         & \hspace{.3cm} {\bf SQ$_1$}: When did Henry Cavill play a fictional character?\\
         & \hspace{.3cm} {\bf SQ$_2$}: Who are the American superhero appearing in American comic books?\\
         \midrule
         \textbf{CLEVR} & {\bf Q1}: How many cubes are small brown objects or rubber things?\\
         & \hspace{.3cm} {\bf SQ$_1$}: How many cubes are small? \\
         & \hspace{.3cm} {\bf SQ$_2$}: What are brown objects or rubber things?\\
         & {\bf Q2}: What material is the small ball that is in front of the big metal cylinder behind\\
         & \hspace{.8cm} the block that is to the left of the small yellow rubber sphere made of?\\
         & \hspace{.3cm} {\bf SQ$_1$}: What material is the small ball?\\
         & \hspace{.3cm} {\bf SQ$_2$}: The big metal cylinder behind the big metal cylinder is\\
         & \hspace{1.3cm} to the left of the small yellow rubber sphere made of?\\
         & {\bf Q3}: There is a object in front of the large cyan rubber thing; what is its material?\\
         & \hspace{.3cm} {\bf SQ$_1$}: Why is there a object in front of the large cyan rubber thing?\\
         & \hspace{.3cm} {\bf SQ$_2$}: What is its material?\\
         & {\bf Q4}: Are there any other things that have the same material as the yellow thing?\\
         & \hspace{.3cm} {\bf SQ$_1$}: Where are there any other things that have the same material?\\
         & \hspace{.3cm} {\bf SQ$_2$}: The yellow thing?\\
         \midrule
         \textbf{Complex} & {\bf Q1}: What is the major religions in UK that believes in the deities ``Telangana Talli''?\\
         \textbf{Web} & \hspace{.3cm} {\bf SQ$_1$}: What is the major religions in UK?\\
         \textbf{Questions} & \hspace{.3cm} {\bf SQ$_2$}: Who believes in the deities ``Telangana Talli''?\\
         & {\bf Q2}: Where to visit in Barcelona that was built before 1900?\\
         & \hspace{.3cm} {\bf SQ$_1$}: Where to visit in Barcelona?\\
         & \hspace{.3cm} {\bf SQ$_2$}: What was built before 1900?\\
         & {\bf Q3}: The person who wrote the lyrics for ``Dirge for Two Veterans'' was influenced by what?\\
         & \hspace{.3cm} {\bf SQ$_1$}: The person who wrote the lyrics?\\
         & \hspace{.3cm} {\bf SQ$_2$}: What was the influence of ``Dirge for Two Veterans''?\\
         & {\bf Q4}: What country with Zonguldak province as its second division speaks Arabic?\\
         & \hspace{.3cm} {\bf SQ$_1$}: What country with Zonguldak province as its second division?\\
         & \hspace{.3cm} {\bf SQ$_2$}: Who speaks Arabic?\\
    \end{tabular}
    \caption{\textbf{Zero-shot Unsupervised Decompositions} of questions or claims from other datasets using our \useqtoseq{} model trained on \hotpot{} and Common Crawl questions (without further, dataset-specific fine-tuning).
    }
  \label{tab:example_decompositions_other_datasets_zeroshot}
\end{table*}

As shown in Table~\ref{tab:example_decompositions_other_datasets_zeroshot}, we decompose queries from several other datasets, using our decomposition model trained on only questions in \hotpot and Common Crawl. In particular, we generate sub-questions for (1) questions in ComplexWebQuestions~\cite{talmor2018web}, which are multi-hop questions about knowledge-bases, (2) questions in CLEVR~\cite{johnson2017clevr}, which are multi-hop questions about images, and (3) claims (statements) in fact-verification challenges, FEVER 1.0~\cite{thorne2018fever} and 2.0~\cite{thorne2019fever2}. These queries differ significantly from questions in \hotpot{} in topic, syntactic structure, and/or modality being asked about. Despite such differences, our trained \useqtoseq{} model often (though not always) generates reasonable sub-questions without any further finetuning, providing further evidence of the general nature of our approach and potential for applicability to other domains.

\section{Single-hop QA Model}
\label{sec:apdx:Single-hop QA Model}
To train the single-hop QA model, we largely follow~\citet{min2019multi} as described below.
We use an ensemble of two models trained on \squad{} 2 and examples from \hotpot{} labeled as ``easy'' (single-hop).
\squad{} is a single-paragraph QA task, so we adapt it to the multi-paragraph setting by retrieving and appending distractor paragraphs from Wikipedia for each question.
We use the TFIDF retriever from DrQA~\cite{chen2017reading} to retrieve two distractor paragraphs, which we add to the input for one model in the ensemble.
We drop words from the question with a 5\% probability to help the model handle any ill-formed sub-questions.

\section{Recomposition Model}
\subsection{Varying Training Set Size}
\begin{figure}[t]
\centering
\includegraphics[width=\columnwidth]{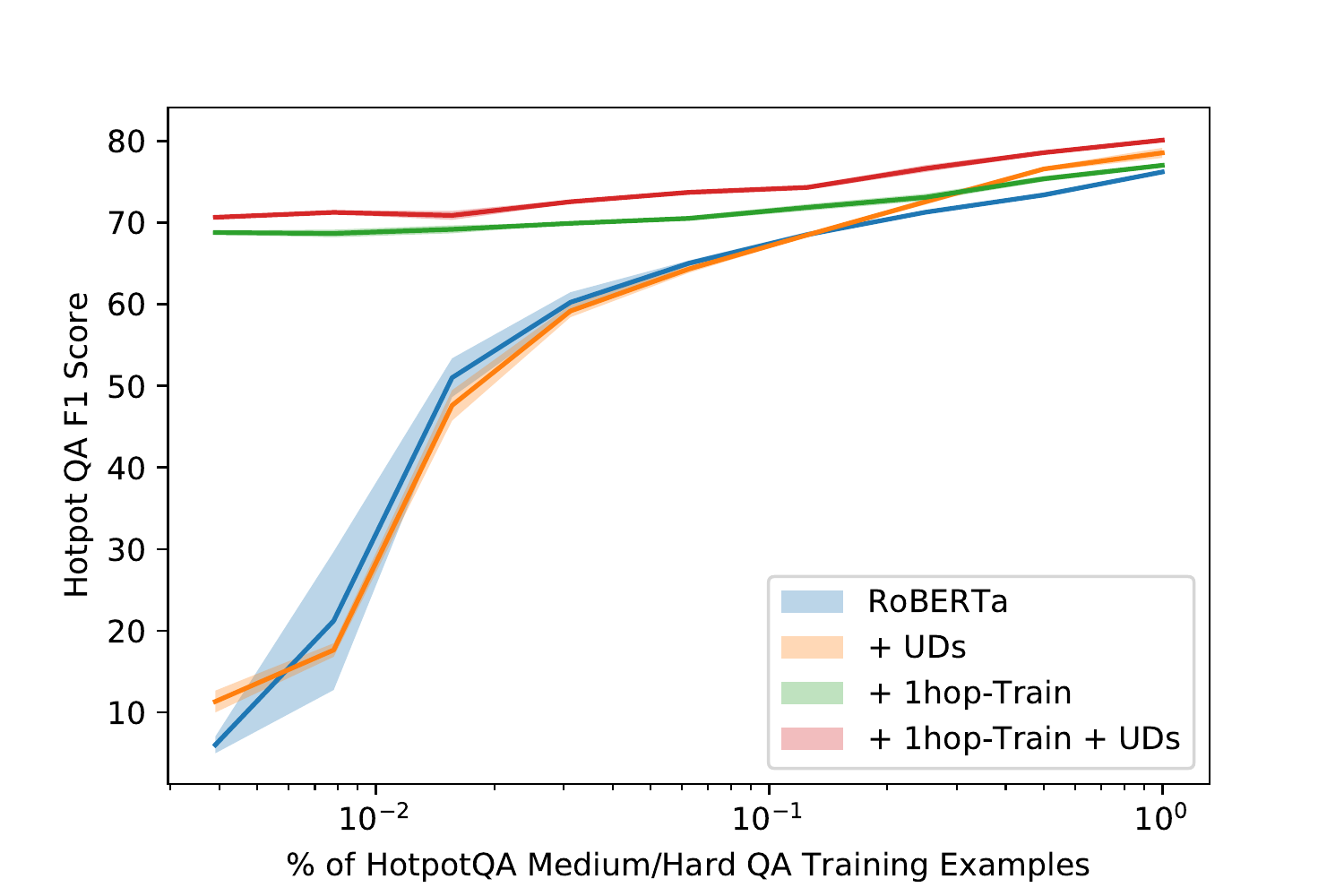}
\caption{QA F1 of the downstream, recomposition model, with and without unsupervised decompositions (UDs), when varying the amount of training data. We also assess the impact of removing single-hop training data (\squad{} 2.0 and \hotpot ``easy'' questions).}
\label{fig:Sample_Efficiency_with_UDs}
\end{figure}

To understand how decompositions impact performance given different amounts of QA training data, we vary the number of multi-hop training examples.
We use the ``medium'' and ``hard'' level labels in \hotpot{} to determine which examples are multi-hop.
We consider training setups where the recomposition model does or does not use data augmentation via training on hotpot{} ``easy''/single-hop questions and \squad~2 questions.
Fig.~\ref{fig:Sample_Efficiency_with_UDs} shows the results.
Decompositions improve QA, so long as the recomposition model has enough training data to achieve a minimum level of performance (here, roughly 68 F1).

\subsection{Improvements across Question Types}
\begin{figure}[t!]
\centering
\includegraphics[width=\columnwidth]{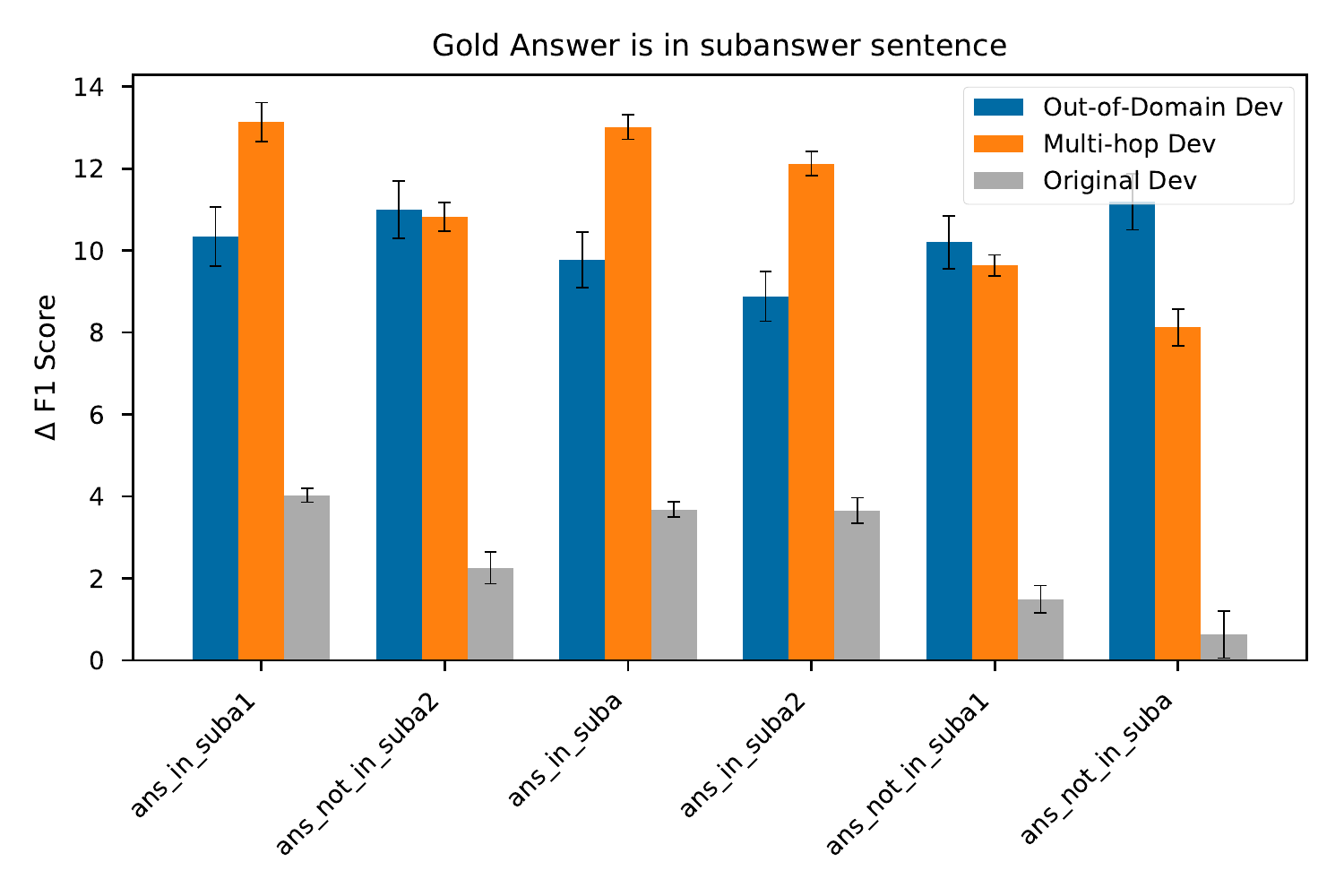}
\caption{Performance difference between a QA model that does vs. does not use \useqtoseq{} decompositions, stratified by whether the gold final answer is in a sub-answer sentence. We find a larger gain when the sub-answer sentence contains the gold, final answer.}
\label{fig:SubAnswer_is_Gold_Answer_bars}
\end{figure}

To better understand where decompositions improve QA, we examined the improvement over the baseline across various fine-grained splits of our three evaluation sets.
Decompositions were roughly as helpful for \yes/\no~questions as for questions with a span-based answer.
Across our evaluation sets, we did not find a consistent pattern regarding what questions, stratified by ``wh-'' question-starting words, benefited the most from decompositions.
Intuitively, we found larger QA improvements when using decompositions when a sub-answer sentence contained a gold, final answer, as shown in Figure~\ref{fig:SubAnswer_is_Gold_Answer_bars}.

\subsection{Training Hyperparameters}
\label{ssec:apdx:multi_hop_qa_training_hyperparameters}
To train \robertalarge, we fix the number of training epochs to 2, as training longer did not help.
We sweep over batch size $\in \{64, 128\}$, learning rate $\in \{1 \times 10^{-5}, 1.5 \times 10^{-5}, 2 \times 10^{-5}, 3 \times 10^{-5}\}$, and weight decay $\in \{0, 0.1, 0.01, 0.001\}$, similar to the ranges used in the original paper~\cite{liu2019roberta}.
We chose the hyperparameters that did best for the baseline QA model (without decompositions) on our dev set: batch size $64$, learning rate $1.5 \times 10^{-5}$, and weight decay $0.01$.
Similarly, for \bert{} experiments, we fix the number of epochs to 2 and choose hyperparameters by sweeping over the recommended ranges from~\citet{devlin2019bert} for learning rate ($\{2 \times 10^{-5}, 3 \times 10^{-5}, 5 \times 10^{-5}\}$) and batch size ($\{16, 32\}$).
For \bertbase, we thus choose learning rate $2 \times 10^{-5}$ and batch size $16$, and for \bertlarge, we use the whole-word masking model with learning rate $2 \times 10^{-5}$ and batch size $32$.
\robertalarge{} and \bertlarge{} have 340M parameters, while \bertbase{} has 110M parameters.
We train all QA models with mixed precision floating point arithmetic~\cite{micikevicius2018mixed}, distributing training across 8, 32GB NVIDIA V100 GPUs, lasting roughly 6 hours.

\clearpage
\begin{table*}[th!]
\centering
\scriptsize
\begin{tabular}{p{1.2cm}p{14.5cm}}
\toprule
\multicolumn{2}{l}{ {\bf Q:} What is the name of the singer who's song was released as the lead single from the album ``Confessions''}\\
\multicolumn{2}{l}{ \hspace{.8cm} and that had popular song stuck behind for 8 consecutive weeks?}\\ \midrule
  \multirow{2}{1cm}{Variable \useqtoseq + FastText}
  & {\bf Sub-Q1} What is the name of the singer? \\
  & {\bf Sub-Q2} Who's the song was released as the lead single from the album? \\
  & \underline{{\bf Sub-Q3} What popular song was stuck behind for 8 consecutive weeks?} \\
  \cmidrule(l){1-2}
  \multirow{2}{1cm}{\useqtoseq + FastText}
  & {\bf Sub-Q1} What is the name of the singer? \\
  & {\bf Sub-Q2} What song was released as the lead single from the album ``Confessions,'' and that had popular song stuck behind for 8 consecutive weeks? \\
  \cmidrule(l){1-2}
  \multirow{2}{1cm}{\useqtoseq + Random}
  & {\bf Sub-Q1} What is the name of the singer who's song was released as the lead single from the album ``Confessions? \\
  & {\bf Sub-Q2} Why did popular song stuck behind for 8 consecutive weeks? \\
  \cmidrule(l){1-2}
  \multirow{2}{1cm}{\seqtoseq + FastText}
  & {\bf Sub-Q1} What is the name of the lead singer of the group? \\
  & {\bf Sub-Q2} When was the last time you heard a song called ``on and on'' by a female vocalist? \\
  \cmidrule(l){1-2}
  \multirow{2}{1cm}{\seqtoseq + Random}
  & {\bf Sub-Q1} What is the maximum number of students allowed to take part in the program? \\
  & {\bf Sub-Q2} What is the maximum number of students allowed per year? \\
  \cmidrule(l){1-2}
  \multirow{2}{1cm}{PseudoD + FastText}
  & {\bf Sub-Q1} Which single was released as the album's lead single? \\
  & {\bf Sub-Q2} Who did the song writing, what are some of the stories behind the songs? \\
  \cmidrule(l){1-2}
  \multirow{2}{1cm}{PseudoD + Random}
  & {\bf Sub-Q1} What was sold to foreign ptts? \\
  & {\bf Sub-Q2} What tends to lead to less money? \\
  \cmidrule(l){1-2}
  \multirow{2}{1cm}{DecompRC}
  & {\bf Sub-Q1} Which singer? \\
  & {\bf Sub-Q2} What is the name of alicia keys's song was released as the lead single from the album "Confessions", and that had popular song stuck behind for 8 consecutive weeks? \\
\bottomrule
\end{tabular}
\vskip -0.7em
\caption{Various decomposition methods for the question \textit{``What is the name of the singer who's song was released as the lead single from the album ``Confessions,'' and that had popular song stuck behind for 8 consecutive weeks?''} Here, the Variable-Length \useqtoseq{} model decomposed the question into three subquestions rather than two.}
\label{tab:appendix_decomp_examples_1}
\end{table*}   

\begin{table*}[th!]
\centering
\scriptsize
\begin{tabular}{p{1.2cm}p{14.5cm}}
\toprule
\multicolumn{2}{l}{ {\bf Q:} Are both Coldplay and Pierre Bouvier from the same country?} \\ \midrule
  \multirow{2}{1cm}{Variable \useqtoseq}
  & {\bf Sub-Q1} Who are similar musical artists to Coldplay? \\
  & {\bf Sub-Q2} Where is Pierre Bouvier from? \\
  \cmidrule(l){1-2}
  \multirow{2}{1cm}{\useqtoseq + FastText}
  & {\bf Sub-Q1} Where are Coldplay and Coldplay from? \\
  & {\bf Sub-Q2} What country is Pierre Bouvier from? \\
  \cmidrule(l){1-2}
  \multirow{2}{1cm}{\useqtoseq + Random}
  & {\bf Sub-Q1} Why are both Coldplay and Pierre Bouvier from the same country? \\
  & {\bf Sub-Q2} What is the purpose of a speech? \\
  \cmidrule(l){1-2}
  \multirow{2}{1cm}{\seqtoseq + FastText}
  & {\bf Sub-Q1} Who are similar musical artists to Pierre Bouvier? \\
  & {\bf Sub-Q2} In the same year, Pierre Bouvier, Pierre Bouvier, Pierre Bouvier and Pierre Bouvier? \\
  \cmidrule(l){1-2}
  \multirow{2}{1cm}{\seqtoseq + Random}
  & {\bf Sub-Q1} What is the maximum number of students allowed to take part in the program? \\
  & {\bf Sub-Q2} What is the maximum number of students allowed to take part in the course of the course of the course of the course of the course of the course of the course of the course of the course of the course of the course of the course of the course of the course of? \\
  \cmidrule(l){1-2}
  \multirow{2}{1cm}{PseudoD + FastText}
  & {\bf Sub-Q1} What do a handful of couples have in common with Coldplay, Coldplay, and Pierre Bouvier? \\
  & {\bf Sub-Q2} Where are the french alps? \\
  \cmidrule(l){1-2}
  \multirow{2}{1cm}{PseudoD + Random}
  & {\bf Sub-Q1} In what year did Fermat declare Fermat's little theorem? \\
  & {\bf Sub-Q2} When did the United States withdraw from the Bretton Woods Accord? \\
  \cmidrule(l){1-2}
  \multirow{2}{1cm}{DecompRC}
  & {\bf Sub-Q1} Is Coldplay from which country? \\
  & {\bf Sub-Q2} Is Pierre Bouvier from which country? \\
\bottomrule
\end{tabular}
\vskip -0.7em
\caption{Various decomposition methods for: \textit{``Are both Coldplay and Pierre Bouvier from the same country?''}}
\label{tab:appendix_decomp_examples_2}
\end{table*}      

\begin{table*}[th!]
\centering
\scriptsize
\begin{tabular}{p{1.2cm}p{14.5cm}}
\toprule
\multicolumn{2}{l}{ {\bf Q:} Who is older, Annie Morton or Terry Richardson?} \\ \midrule
  \multirow{2}{1cm}{Variable \useqtoseq}
  & {\bf Sub-Q1} Who is Annie Morton? \\
  & {\bf Sub-Q2} Who was Terry Morton? \\
  \cmidrule(l){1-2}
  \multirow{2}{1cm}{\useqtoseq + FastText}
  & {\bf Sub-Q1} Who is Annie Morton? \\
  & {\bf Sub-Q2} When was Terry Richardson born? \\
  \cmidrule(l){1-2}
  \multirow{2}{1cm}{\useqtoseq + Random}
  & {\bf Sub-Q1} Who is older, Annie Morton? \\
  & {\bf Sub-Q2} Who is Terry Richardson? \\
  \cmidrule(l){1-2}
  \multirow{2}{1cm}{\seqtoseq + FastText}
  & {\bf Sub-Q1} Who is Terry Morton? \\
  & {\bf Sub-Q2} Who is Terry Morton? \\
  \cmidrule(l){1-2}
  \multirow{2}{1cm}{\seqtoseq + Random}
  & {\bf Sub-Q1} What is the maximum number of students allowed to take part in the program? \\
  & {\bf Sub-Q2} What is the maximum number of students allowed to take part in the program? \\
  \cmidrule(l){1-2}
  \multirow{2}{1cm}{PseudoD + FastText}
  & {\bf Sub-Q1} Who was Terry Richardson? \\
  & {\bf Sub-Q2} Who changes positions more, Annie Morton, Cotton Hill, or Terry Richardson? \\
  \cmidrule(l){1-2}
  \multirow{2}{1cm}{PseudoD + Random}
  & {\bf Sub-Q1} What did Decnet Phase I become? \\
  & {\bf Sub-Q2} What group can amend the Victorian constitution? \\
  \cmidrule(l){1-2}
  \multirow{2}{1cm}{DecompRC}
  & {\bf Sub-Q1} Annie Morton is born when? \\
  & {\bf Sub-Q2} Terry Richardson is born when? \\
\bottomrule
\end{tabular}
\vskip -0.7em
\caption{Various decomposition methods for: \textit{``Who is older, Annie Morton or Terry Richardson?''}}
\label{tab:appendix_decomp_examples_3}
\end{table*}

\end{document}